\documentclass[review]{elsarticle}

\usepackage{lineno,hyperref}
\usepackage{subfig}
\usepackage{graphicx}
\usepackage{multirow}
\usepackage[table,xcdraw]{xcolor}
\usepackage{amsmath,amssymb}
\usepackage{amsmath}
\modulolinenumbers[5]

\journal{Elsevier Latex Template}

\begin{document}

\begin{frontmatter}

\title{Fast \& Slow Learning: Incorporating Synthetic Gradients in Neural Memory Controllers}

\author[label1]{Tharindu Fernando\corref{cor1}}
\ead{t.warnakulasuriya@qut.edu.au}
 
\author[label1]{Simon~Denman}
\ead{s.denman@qut.edu.au}
 
\author[label1]{Sridha Sridharan}
\ead{s.sridharan@qut.edu.au}
 
\author[label1]{Clinton~Fookes}
\ead{c.fookes@qut.edu.au}

\cortext[cor1]{Corresponding author at: Signal Processing, Artificial Intelligence and Vision Technologies (SAIVT), Queensland University of Technology, Australia.}
\address[label1]{Signal Processing, Artificial Intelligence and Vision Technologies (SAIVT), Queensland University of Technology, Australia.}

\begin{abstract}
Neural Memory Networks (NMNs) have received increased attention in recent years compared to deep architectures that use a constrained memory. Despite their new appeal, the success of NMNs hinges on the ability of the gradient-based optimiser to perform incremental training of the NMN controllers, determining how to leverage their high capacity for knowledge retrieval. This means that while excellent performance can be achieved when the training data is consistent and well distributed, rare data samples are hard to learn from as the controllers fail to incorporate them effectively during model training. Drawing inspiration from the human cognition process, in particular the utilisation of neuromodulators in the human brain, we propose to decouple the learning process of the NMN controllers to allow them to achieve flexible, rapid adaptation in the presence of new information. This trait is highly beneficial for meta-learning tasks where the memory controllers must quickly grasp abstract concepts in the target domain, and adapt stored knowledge. This allows the NMN controllers to quickly determine which memories are to be retained and which are to be erased, and swiftly adapt their strategy to the new task at hand. Through both quantitative and qualitative evaluations on multiple public benchmarks, including classification and regression tasks, we demonstrate the utility of the proposed approach. Our evaluations not only highlight the ability of the proposed NMN architecture to outperform the current state-of-the-art methods, but also provide insights on how the proposed augmentations help achieve such superior results. In addition, we demonstrate the practical implications of the proposed learning strategy, where the feedback path can be shared among multiple neural memory networks as a mechanism for knowledge sharing. 
\end{abstract}


\end{frontmatter}


\section{Introduction}

{D}{eep} learning has achieved tremendous success in numerous large-scale supervised learning tasks, including object tracking \cite{zhai2018deep}, behaviour recognition \cite{gammulle2019coupled,gammulle2020fine} and prediction \cite{gupta2018social}, and games \cite{silver2017mastering}. In particular, Neural Memory Networks (NMNs) have achieved a notable improvement \cite{yang2018learning,parisotto2017neural,fernando2018learning,fernando2019memory,fernando2018pedestrian,fernando2020neural,gammulle2019forecasting} compared to deep architectures that use a constrained memory, as a result of NMNs capacity to map dependencies over long-time durations. Their success hinges on the ability of the gradient-based optimiser to perform incremental training, leveraging the high model capacity \cite{santoro2016meta}. However, most deep learned architectures fail in one-shot and few-shot learning settings, where they are required to rapidly generate inferences from limited data. The one or few-shot learning task generally falls within the field of meta-learning \cite{thrun1998lifelong, vilalta2002perspective, liu2020meta, xu2020meta, li2020revisiting}, where the learning happens in two stages. In the first stage, rapid learning occurs within a particular task, such as learning to classify images in a source domain. Then the acquired knowledge is transferred to the target domain which has a different task structure \cite{santoro2016meta}.  

Prior works \cite{hochreiter2001learning, santoro2016meta} have shown the potential of NMNs, which are equipped with external memory slots and can be leveraged to rapidly cache new representations to the memory for meta-learning. NMNs are a variant of Recurrent Neural Networks where an external memory stack is used to store informative information, instead of an internal cell state which is typically used by Long Short-Term Memory (LSTM) \cite{hochreiter1997long} Networks and Gated Recurrent Units (GRUs) \cite{chung2014empirical}. The issue with LSTMs and GRUs is the fact that the content is erased when a new sequence comes in \cite{fernando2018tree, fernando2018learning}. This is because such architectures are designed to map temporal relationships within a sequence, not in between sequences \cite{fernando2018tree, liang2016semantic, liang2017interpretable, gammulle2019forecasting}. Hence, the limited capacity of the internal cell state is not sufficient to model relationships in the entire dataset \cite{ma2019taxonomy, fernando2018tree}. The NMNs leverage the additional capacity of the external memory stack to model the informative characteristics among the different data samples in the dataset. As external memory is relatively new concept, we refer interested readers to \cite{ma2019taxonomy} where the authors illustrate the taxonomy of NMNs and their general architectural details.

It is apparent that the success of these NMN techniques to successfully accomplish meta-learning tasks depends upon the ability of the memory controllers to quickly grasp the abstract concepts in the target domain, and adapt the stored knowledge as required. 

\begin{figure*}[htbp]
    \centering
    \subfloat[][]{\includegraphics[width=.55\linewidth]{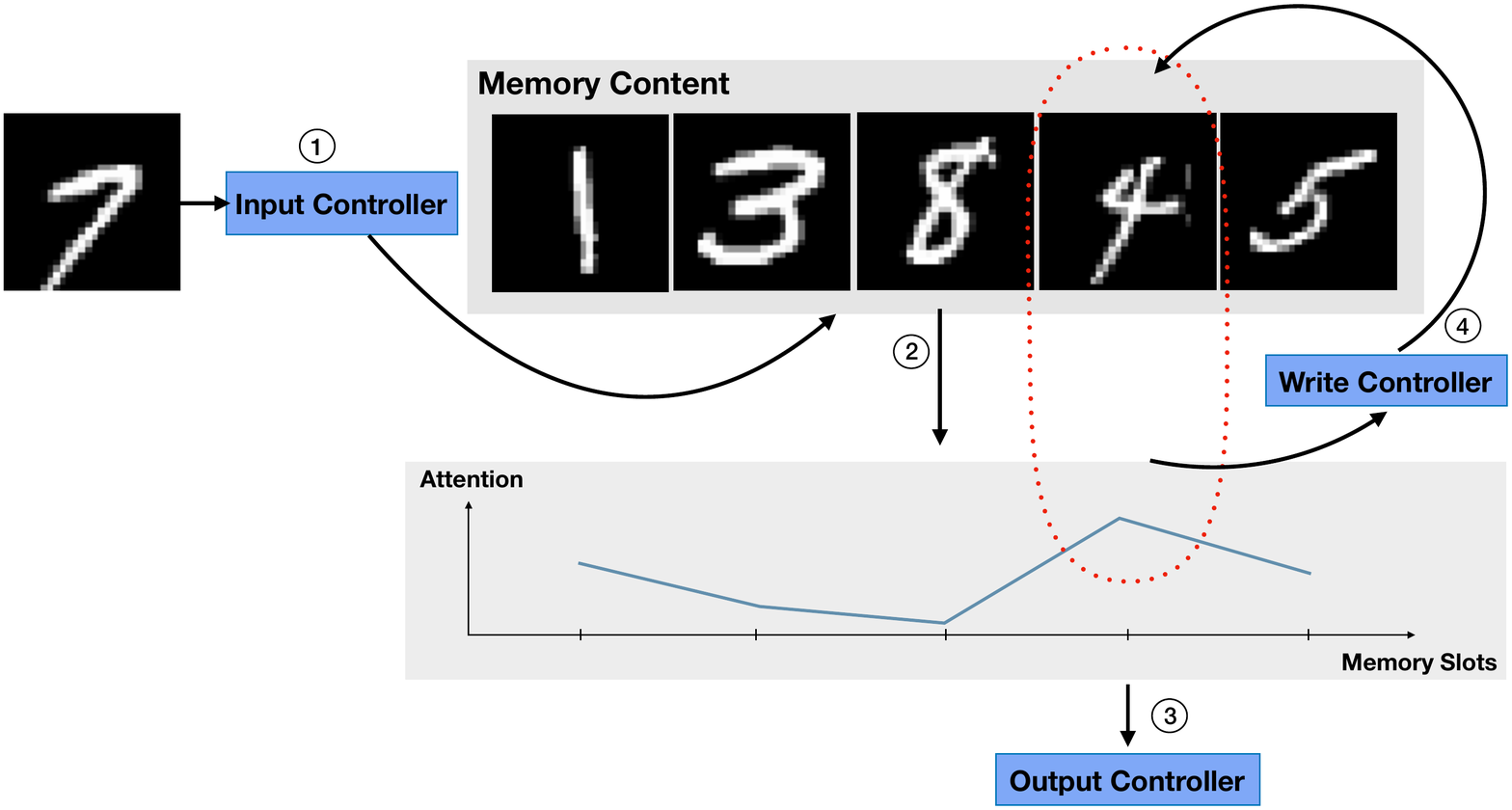}}
    \subfloat[][]{\includegraphics[width=.55\linewidth]{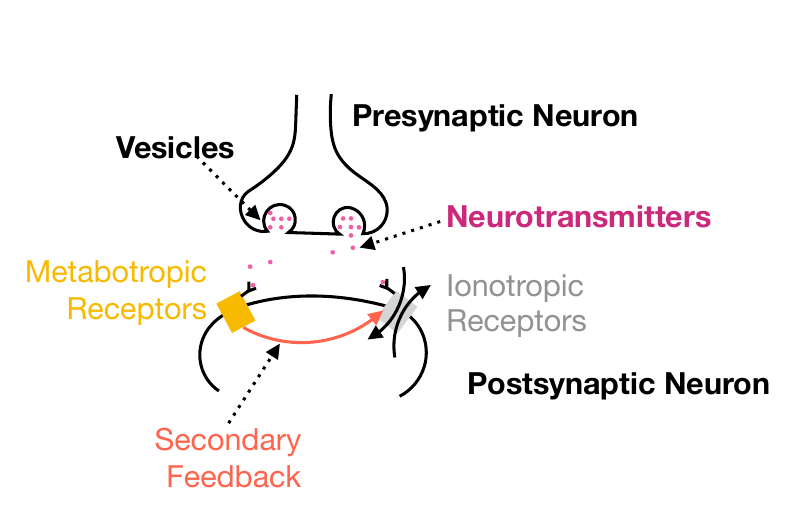}}\\
    \caption{Motivation: (a) A simplified figure illustrating the deficiencies of the current neural memory network controller structure. The memory has never seen the digit 7, however, the output controller gives more attention to digit 4 considering it's spatial similarities to 7. Hence, the write controller also partially updates the digit 4 memory embedding using information from digit 7, without completely incorporating digit 7 within the memory. (b) Functionality of neuromodulators in the human learning process. Metabotropic receptors act as a secondary feedback path to the Ionotropic receptors, dynamically changing the transmission between neurons, resulting in fast or slow neurocircuitry. }
    \label{fig:challenge}
\end{figure*}

A typical memory module is composed of a memory stack which stores the information, and a set of controllers (an input controller, an output controller and a write controller) to control the module's functionality. Following the current optimisation strategy applied to NMNs, the prediction error is back propagated to update the behaviour of the controllers. However, this strategy is inadequate in the presence of new and diverse observations (from the target domain) which are highly fragile. Fig. \ref{fig:challenge} (a) illustrates the deficiencies of the existing NMN training process. In this simplified representation, the current state of the memory is composed of MNIST digits 1, 3, 8, 4, 5, and the model has never seen the digit 7 which is the current input. The input controller of the NMN encodes the input and then quantifies the similarity between the input and the content of each memory slot, shown via the activation in the figure (note that this process is discussed in more detail in Sec. \ref{sec:basic_architecture}, however this simplified example illustrates the general operational structure).   

Even if the model has never seen an example of digit 7, the output controller tries to formulate an output using the existing memory content, hence, more attention is given to digit 4, considering its spatial similarities to digit 7. A second errornous operation takes place when the write controller updates the memory by updating the content of the memory slot which corresponds to digit 4, as it contributed more (i.e more attention) when generating the output. However, what we'd like the write controller to do is identify that digit 7 is an example that it has never seen, and add it to the memory content instead of updating the representation of digit 4. Yet, the current learning strategy is directly coupled with examples that contribute to the majority of the model's loss, hence, if digit 7 is a rare observation the information that was written when it was seen will be simply overridden and discarded. We observe that these errornous operations occur due to the inability of the controllers to identify such rare information and quickly learn from it. A naive solution to this problem would be to augment the output and write controller strategies to use a similarity score, and change the operations based on whether the similarity between the memory content and the input is greater than a threshold, adding new content if the similarity is less than the threshold. However, such an operation is sub-optimal due to memory size constraints, and if the memory size is limited then there still exists the tendency to overwrite less frequent examples as they have less impact on the overall error.


In contrast, a greater flexibility for managing new information is celebrated in human learning, where humans demonstrate superior ability to solve a new task that they have never seen before. Hence, we draw our inspiration from the human learning process. We observe that the brain utilises multiple neuromodulators to determine which memories are to be retained and which are to be erased \cite{marblestone2016toward}.  Fig. \ref{fig:challenge} (b) visually illustrates this. Specifically, the Metabotropic receptors act as a secondary feedback path to the Ionotropic receptors and control their behaviour, as well as changing the activity of the proteins within the postsynaptic neuron. The functionality of neuromodulators, which dramatically transforms the functional output of neural circuitry \cite{bargmann2012beyond}, is controlled by the hippocampus of the human brain. The hippocampus determines whether learning must occur quickly on the basis of a single episode, while in other contexts the learning will happen slowly by aggregating patterns across large amounts of data \cite{marblestone2016toward} (hence the term fast and slow learning). This clearly indicates the need for such a mechanism in NMNs to generate fast and dynamic responses, in addition to the usual slow responses provided by gradient descent. In the presence of less frequent or rare information, if there is a mechanism to facilitate additional feedback to the NMN controllers such that they can alter their operations to preserve important information, then such deficiencies can be fully rectified. We observe the feasibility of formulating such feedback structure using the Synthetic Gradient (SG) framework of \cite{jaderberg2017decoupled}. It should be noted that we do not seek to fully replicate the neuromodulation process using SGs. We only use the notion of neuromodulation to formulate the concept of fast and slow learning, and the existence of multiple feedback paths. 

In \cite{jaderberg2017decoupled} the authors propose an alternative to back propagation in which they add a secondary feedback path where the neurons of a layer are updated using local layer-wise predictors, which approximate the loss of that particular layer. This allows arbitrary feedback paths among different layers, and removes the constraint that the layers of a particular network need to be temporally locked. While drawing inspiration from this SG framework of \cite{jaderberg2017decoupled}, we propose to decouple the learning process of the NMN controllers such that they can demonstrate flexible, rapid adaptation in the presence of new information. To the best of our knowledge this is the first work which investigates the augmentation of NMN controller behaviours to facilitate meta-learning. With qualitative evaluations we show that Synthetic Gradients formulate a fast and slow learning paradigm where fast learning happens in the presence of new and rare information while slow learning takes place with more frequent observations. The main contributions of the proposed work can be summarised as follows:
\begin{itemize}
    \item Inspired by human cognition, we propose a novel optimisation strategy to learn the controller behaviour for NMNs. 
    \item We formulate multiple feedback paths to facilitate dual rapid and gradual learning processes of the controllers, enabling effective learning using both new and rare information, and more frequent observations. 
    \item We provide empirical evaluations on multiple public benchmarks which indicate the utility of the proposed approach.
    \item We illustrate practical implications of the proposed learning strategy, where the feedback path can be shared among multiple neural memory networks. 
\end{itemize}

\section{Related Works}
The task that this paper addresses can be broadly categorised into the problem of meta-learning. One widely used approach for meta-learning is to adapt the learner's (i.e the main model's) parameters through a secondary model (i.e the meta-learner) \cite{bengio1990learning, bengio1992optimization, schmidhuber1992learning}. With the dawn of deep learning this approach has been extensively applied for learning to optimise deep neural networks, \cite{andrychowicz2016learning}, for initialising and optimising few-shot image recognition models \cite{ravi2016optimization}, to generate weights of other networks \cite{ha2016hypernetworks}, and for neural architecture search \cite{zoph2016neural}.  

One of the most notable approaches among these methods is Model-agnostic meta-learning for fast adaptation of deep networks (MAML) \cite{finn2017model}, where the meta-knowledge is utilised for the initialisation of model parameters where the fewest number of training instances are required to adapt the learned model to a new task. This architecture was further extended in \cite{antoniou2018train} where the authors introduce numerous augmentations to further stabilise and generalise the MAML architecture. 

Unlike these methods, where an explicit change in model parameters is used to adapt the model to the new data, inspired by the biological cognition process we investigate how the stored embedding in an external memory can be adapted to transform the model for the new task. 

Our proposed architecture is also distinct to existing memory-augmented meta-learning architectures. Specifically, the meta-NMN architecture proposed in \cite{santoro2016meta} utilises general LSTM controllers for memory read and output operations, and the authors have proposed to augment the memory write operation by incorporating the least recently used access module for the memory write operation. In \cite{munkhdalai2017meta} the authors propose to utilise a combination of fast and slow weights. The fast weights are a set of predicted weights per task obtained by the meta-network, while the slow weights are weights from the REINFORCE algorithm \cite{williams1992simple} trained across all the tasks. Here an external memory is utilised to store the network parameters. 

In contrast, the proposed approach investigates augmenting the behaviour of the controllers of the NMN such that the memory read, compose and write controllers can dynamically change their behaviour and demonstrate fast and slow learning capabilities. This allows the NMN to quickly identify salient information relevant to the new task at hand, and efficiently store and retrieve this information. We would like to point out the parallels between the proposed approach and the adaptation of the human cognition process using neuromodulators, where the hippocampus identifies the salient information that should be quickly comprehended while allowing other information to be more slowly grasped. Most importantly, it is observed that neuromodulators in human brain alter the functional connectivity of information flow in different time intervals, allowing comprehension at fast and slow timescales \cite{bargmann2012beyond}. Even though SGs in the proposed method do not directly mimic the functionality of the neuromodulators, we would like to highlight the similarities as mechanisms facilitating the fast-slow learning in memory modules while modulating their operations. 

The proposed work also relates to the concept of synthetic gradients introduced in \cite{jaderberg2017decoupled}, which demonstrated that learning can be based on approximate gradients. This method has been recently applied to distributed training across cloud and edge devices. Specifically, the authors in \cite{chen2019exploring} model parallel training of a deep neural network (DNN) model, which decouples the model layer-wise and allows the different layers to be hosted by different, resource-constrained devices. In a different line of work \cite{lansdell2019learning} propose a network that learns to use feedback signals trained using reinforcement learning via a global reward signal; while the authors in \cite{shang2019alternating} explore ways of combining synthetic and real gradients with applications to neural language modeling tasks.

In addition, in \cite{czarnecki2017understanding} the authors have conducted further analysis with respect to the characteristics of the SG-based models and demonstrated that they converge faster compared to true gradient based methods. However, these analyses were conducted with simple neural network architectures and to the best of our knowledge no prior work has either investigated its effectiveness in a meta-learning setting, or its potential to augment the behaviour of NMN controllers. However, we would like to point-out that this is the first work which investigates the application of SG to augment the functionality of NMN controllers. Even though the authors of \cite{jaderberg2017decoupled} have demonstrated the application of SG in recurrent architectures using RNNs, RNNs have a significantly different architecture compared to NMN controllers. Firstly, the RNNs map the relationships within a single example while NMNs model the relationships in between examples (i.e within the entire dataset) \cite{fernando2020neural, fernando2019memory}. Secondly, when the authors of \cite{jaderberg2017decoupled} decouple the learning process of RNNs using SGs, the SGs regulate the output of the RNN itself, however, the NMN controllers have a broader functionality as they control the content of a memory stack. In addition, in the proposed work we are modulating the behaviour of three components using three separate SG modules. Despite their functional separation, the controllers should collectively operate to generate informative outputs using the memory. Hence, this work exhibits a more challenging and distinct functionality of SGs. Furthermore, to the best of our knowledge none of the previous works have investigated the utilisation of SGs in the context of meta learning. With experimental evaluations we illustrate the behavioural utility of SGs to facilitate fast learning in the presence of new and rare information, which is a highly beneficial characteristic in meta-learning.    

\section{Architecture}
This section first introduces the structure of a typical NMN, the controller modules and their functionality. Then we illustrate the proposed procedure to decouple the learning of the controllers through synthetic gradients.

\subsection{Neural Memory Networks}
\label{sec:basic_architecture}
\begin{figure}[htbp]
    \centering
    \includegraphics[width=.85\linewidth]{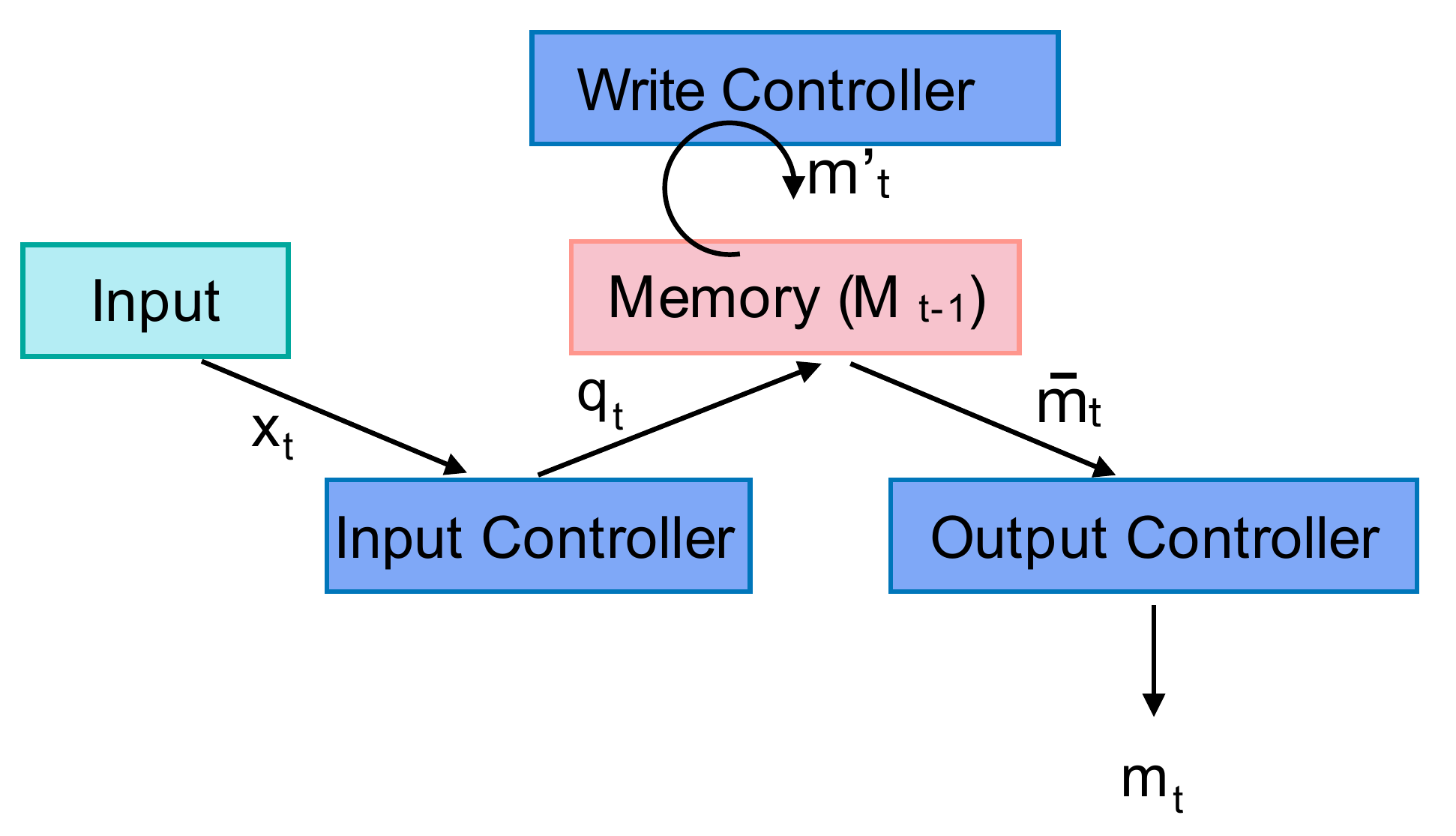}
    \caption{Overview of an external memory which is composed of input, output and write controllers. The input controller determines what facts within the input are used to query the memory. The output controller determines what portion of the memory is passed as the output for that query. Finally the write controller updates the memory state and propagates it to the next time step.}
    \label{fig:basic_nmn}
\end{figure}

Fig. \ref{fig:basic_nmn} shows the composition of a typical memory module which is composed of, (1) a memory stack which stores the information, (2) a input controller which queries the information stored in the memory, (3) an output controller to control the information passed out from the memory, and (4) a write controller to update the memory and evolve it temporally. 

Formally, we define the memory stack as $M \in \mathbb{R}^{l \times k}$ which contains $l$ memory slots, each with an embedding dimension of $k$. The state of the memory at time instance $t-1$ is given by $M_{t-1}$. We first pass the input, $\chi_t$, at the current time instance $t$, through a feature extractor function, $f^E$, and generate a feature vector, $x_t$, such that,
\begin{equation}
x_t = f^{E} (\chi_t).
\end{equation}

The input controller in the memory module receives $x_t$ and it is passed through a read function, $f_r^{LSTM}$, which is composed of LSTM cells, and generates a query vector, $q_t$, such that, 
\begin{equation}
q_t = f^{LSTM}_r (x_t).
\end{equation}

Motivated by \cite{munkhdalai2017neural,fernando2020neural}, a softmax function is used to quantify the similarity between each memory slot in $M_{t-1}$ and the query vector, $q_t$,
\begin{equation}
z_t = \textrm{softmax}({q_t}^\top M_{t-1}).
\end{equation}

Using this score vector, $z_t$, the output controller generates the memory output, $m_t$, such that,
\begin{equation}
 \Bar{m}_t=z_{t} [M_{t-1}]^\top,
\end{equation}
and, 
\begin{equation}
 m_t=f^{LSTM}_o(\Bar{m}_t).
\label{eq:memory_out}
\end{equation}

As the final step the write controller generates an update vector, $m'_t$, to update the memory,
\begin{equation}
m'_t = f_w^{LSTM} (m_t),
\end{equation}
and updates the memory using,
\begin{equation}
M_t = M_{t-1} (I - z_{t} \otimes e_k)^\top + (m'_t \otimes e_l) (z_{t} \otimes e_k)^\top,
\end{equation}
where $I$ is a matrix of ones, $e_l \in \mathbb{R}^l$ and $e_k \in \mathbb{R}^k$ are vectors of ones and $\otimes$ denotes the outer product which duplicates its left vector $l$ or $k$ times to form a matrix. 

Memory output, $m_t$, is then utilised by the decoder component of the network, which is parameterised by the function, $f^{D}$, which generates the prediction, $\hat{y}_t$, such that,
\begin{equation}
\hat{y}_t = f^{D}(m_t).
\end{equation}

\subsection{Injecting a Secondary Feedback Path to Controllers}
\label{sec:sgs}
\begin{figure*}[htbp]
    \centering
    \includegraphics[width=.95\linewidth]{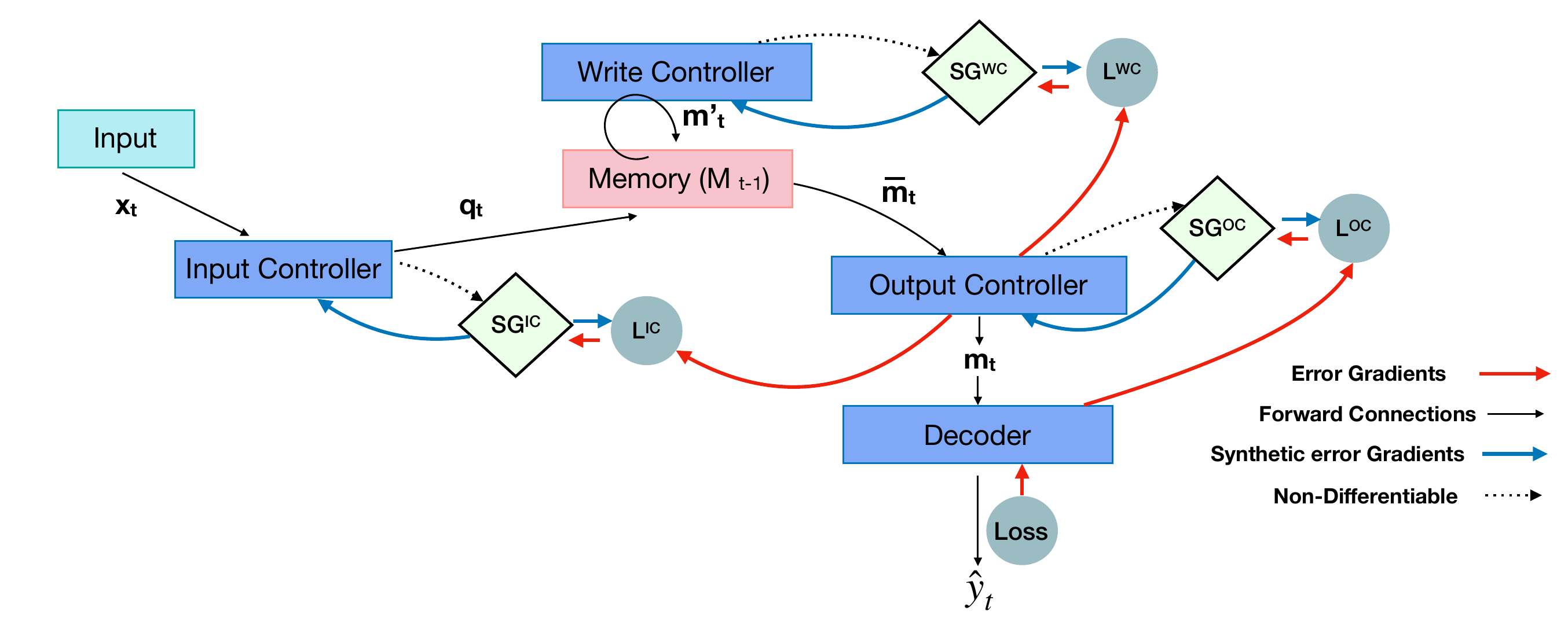}
    \caption{Proposed Framework for Incorporating Synthetic Gradients into the Memory Controllers. The gradient predictor, $SG^{IC}$, of the  Input Controller, $IC$, predicts the loss of $IC$ for its output $q_t$. This additional feedback path is utilised by the $IC$ to dynamically adjust its behaviour to different tasks at hand. Similarly, we define gradient predictors, $SG^{OC}$, and $SG^{WC}$, for output and write controllers respectively.}
    \label{fig:sg_nmn}
\end{figure*}

Motivated by the Synthetic Gradient (SG) introduced in \cite{jaderberg2017decoupled}, we propose to augment the feedback paths of the individual memory controllers using SGs. For instance, as shown in Fig. \ref{fig:sg_nmn} the error gradients of the Input Controller (IC) can be approximated using it's output, $q_t$, and the ground truth output of the decoder network, $y_t$, using the gradient predictor module of the IC, $SG^{IC}$, 

\begin{equation}
    \frac{\delta L^{IC}}{ \delta q_t} = SG^{IC}(q_t, y_t),
\end{equation}

which allows us to update the weights of the IC, $\theta^{IC}$, using a secondary feedback path such that,
\begin{equation}
\theta^{IC} \xleftarrow{} \theta^{IC} - \alpha SG^{IC}(q_t, y_t) \frac{\delta q_t}{ \theta^{IC}},
\end{equation}

where $\alpha$ is the learning rate. It should be noted that the parameters of the $SG^{IC}$ module, $\theta^{SG^{IC}}$, are trained by minimising the $L_{2}$ distance between the predicted and true gradient loss, 
\begin{equation}
L^{SG^{IC}} = || SG^{IC}(q_t, y_t) - \frac{\delta L^{IC}}{ \delta q_t} ||^2.
\end{equation}

Similarly we define SG modules for output and write controllers, $SG^{OC}$, and $SG^{WC}$, respectively. The complete network diagram is given in Fig. \ref{fig:sg_nmn}. The key aspect to note is that the alternative feedback paths that the individual SGs provide to the memory controllers facilitate efficient learning of the controllers, directly observing their impact on the output. This results in an indirect way for the controllers to swiftly adapt their learned behaviour, in spite of the back-propagation of the true gradient. 

\section{Evaluations}

In this section we compare the performance of the proposed method with prior meta-learning algorithms on the popular few-shot learning benchmark, the Omniglot \cite{lake2011one} dataset, as well as the real world regression task of trajectory prediction on the inD \cite{bock2019ind} dataset. 

\subsection{Handwritten Character Classification on the Omniglot Dataset}
\label{sec:classification}
The Omniglot dataset consists of 1,623 characters extracted from 50 different alphabets, where each of these instances was hand drawn by a different person. We follow the evaluation protocol outlined in \cite{vinyals2016matching} where an N-way classification task is setup as follows: randomly pick $N$ classes from the data and provide the model with $S$ samples from each of the $N$ classes, and evaluate the model using new samples within the $N$ classes. Following the data augmentation approach of \cite{finn2017model}, we applied random rotations by multiples of 90 degrees. We used 1200 characters for training and the remainder for testing. We evaluated both the 5-way and 20-way (i.e $N=5$ and $N=20$) classification settings, using both 1-shot and 5-shot classification. The input images are down-sampled to $28 \times 28$ pixels.

For the feature extractor network, $f^{E}$, we used 3 modules with $3 \times 3$ convolutions and 64 filters followed by batch normalisation, a ReLu non-linearity and  $2 \times 2$ max pooling. The resultant feature vector is flattened and passed through a memory module with 10 slots. As the decoder function, $f^{D}$, we use a dense layer with softmax activation. For all SGs we utilise LSTM units which accept the respective output vector of the individual controllers, which is concatenated together with the ground truth output vector of the decoder. 

Both frameworks (main framework and the SGs) are optimised using the Adam optimiser with a $5e^{-6}$ learning rate, and for $f^{D}$ we use categorical cross entropy loss. For the SGs we use Mean Square Error (MSE) loss. 

Evaluation results are presented in Tab. \ref{tab:omniglot}. As baselines we report the state-of-the-art MAML \cite{finn2017model}, MANN \cite{santoro2016meta}, MetaNet \cite{munkhdalai2017meta}, and MAML++ \cite{antoniou2018train} models. Furthermore, we report the accuracy for the baseline memory network which has an identical structure except for the utilisation of SGs. 

\begin{table}[htbp]
\centering
\caption{Few-shot classification results on the held-out Omniglot data set. We report accuracies for both the 5-way and 20-way (i.e $N=5$ and $N=20$) classification setting using both 1-shot and 5-shot classification.}
\begin{tabular}{|c|c|c|c|c|}
\hline
\multirow{2}{*}{Model} & \multicolumn{2}{c|}{5-Way} & \multicolumn{2}{c|}{20-Way} \\ \cline{2-5} 
                       & 1-shot       & 5-shot      & 1-shot       & 5-shot       \\ \hline
          baseline NMN &  80.2        &  89.1       & 78.1         & 83.5          \\ \hline
          MANN  \cite{santoro2016meta}       &  82.8        &  94.9       & -            &  -          \\ \hline
          MAML  \cite{finn2017model}       &  98.7        &  \textbf{99.9}       &  95.8        &  98.9        \\ \hline
          MetaNet \cite{munkhdalai2017meta}     &  98.9        &  -          &  97.0        &   -          \\ \hline
          MAML++ \cite{antoniou2018train}      &  -           &   -         &  97.6        &  \textbf{99.3}        \\ \hline \hline
          Proposed     &  \textbf{99.0}        &   99.7      &  \textbf{97.9}        &  99.1        \\ \hline 
          
\end{tabular}
\label{tab:omniglot}
\end{table}

When considering the results presented in Tab. \ref{tab:omniglot} we observe that the proposed method achieves significant performance gains, especially when compared with the memory based meta-learning baselines \cite{santoro2016meta, munkhdalai2017meta}. Furthermore, when compared with the baseline memory architecture we observe that this network fails to capture the semantics of the new task where it is required to quickly adapt the functionality of the controllers. This is a crucial behaviour required for meta-learning as the controllers are the backbone when determining how memory output is retrieved and what information is saved in the memory. In contrast, we observe a significant robustness in the proposed framework in both 5-way and 20-way classification tasks. One particular observation is that the proposed method performs slightly worse than the MAML models \cite{finn2017model} and \cite{antoniou2018train} in the 5 shot classification setting. However, the proposed method outperforms both in the 1-shot classification setting. We believe this is because MAML models specifically try to adapt the learned model to the new tasks. Due to MAML models being light weight models \cite{finn2017model} they can quickly adapt using the samples in the 5-shot evaluation setting. In contrast the memory based models need to adapt the representation of the memory content, requiring more parameter tuning. We believe this causes the difference in performance between the 1-shot and 5-shot classification settings for the proposed and MAML models. However, the proposed method outperforms MAML models in the 1-shot classification setting (in both 5-way and 20-way classification tasks). This is due to use of the memory stack, which capture the semantics in the initial training examples, and thus NMNs can better make sense of the little information available in the 1-shot setting. However, when the fine-tuning happens (between 1-shot to 5-shot) MAML can adapt quicker. Yet, we would like to note the fact that the performance difference is slight and the proposed SG based controller adaptation strategy has been able to more eagerly adapt the controller behaviour compared to rest of the memory-based baselines.  

\subsection{Trajectory Prediction on inD \cite{bock2019ind} Dataset}
\label{sec:regression}
In this evaluation we illustrate a real world application of meta-learning where the prediction models are trained using a sub-group of trajectories, and then they are rapidly tuned to regress another group of trajectories. 

In this evaluation we utilise the inD \cite{bock2019ind} dataset which contains a multitude of different trajectory patterns from 11,500 road users extracted from 4 different German intersections. Hence this dataset has a greater variability among trajectories. Furthermore, in contrast to existing corpora like Stanford Drone \cite{robicquet2016forecasting} and ETH Hotel \cite{yamaguchi2011you}, it contains a greater variety of road user types, including, cars, trucks, buses, pedestrians and cyclists. We formulate the evaluation as follows: We first trained the model using cyclist trajectories where we sample the first 50 frames from the trajectory as the input and the model predicts the trajectory for the next 50 frames. For the feature extraction function, $f^{E}$, we utilise an LSTM with 30 hidden units and the memory module contains 10 memory slots. As the decoder function, $f^D$ , we use a dense layer with 2 units and a relu activation. We use the $X$ and $Y$ Cartesian coordinates of the trajectory as the input sequence, and scale them to the range [-0.5 to 0.5]. 

\begin{figure}[htbp]
    \centering
    \includegraphics[width=.95\linewidth]{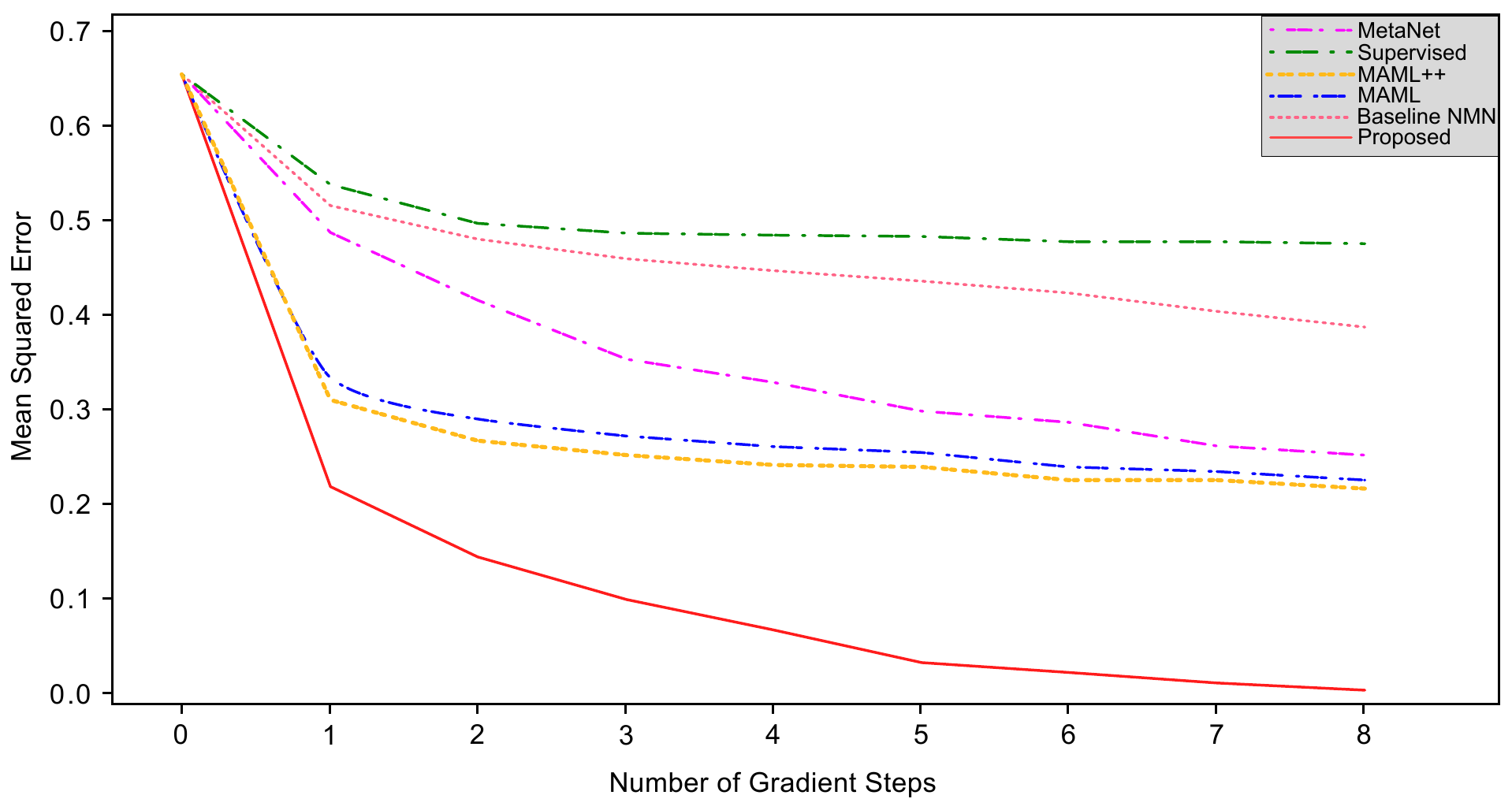}
    \caption{Quantitative trajectory prediction results for the inD dataset showing the learning curve at meta test-time.}
    \label{fig:track_loss}
\end{figure}

We used 150 trajectories for model training and trained both the main framework and the SGs using MSE loss with the Adam optimiser and a learning rate of $1e^{-5}$ for 100 training iterations. After 100 iterations, we sample 10 trajectories randomly from the pedestrian group and report the MSE after each training iteration. Fig. \ref{fig:track_loss} illustrates the learning curve for this meta-testing scenario. As baselines we report the results for the same memory model trained without using the SG predictors in the controllers, MAML \cite{finn2017model}, MetaNet \cite{munkhdalai2017meta}, and MAML++ \cite{antoniou2018train} models. Furthermore, we evaluate a framework which is specifically designed for trajectory prediction \cite{fernando2018gd}. This model is pre-trained using the same training setting and fine-tuned with pedestrian trajectories using the same learning rate. 

\begin{figure*}[htbp]
    \centering
    \subfloat[][]{\includegraphics[width=.45\linewidth]{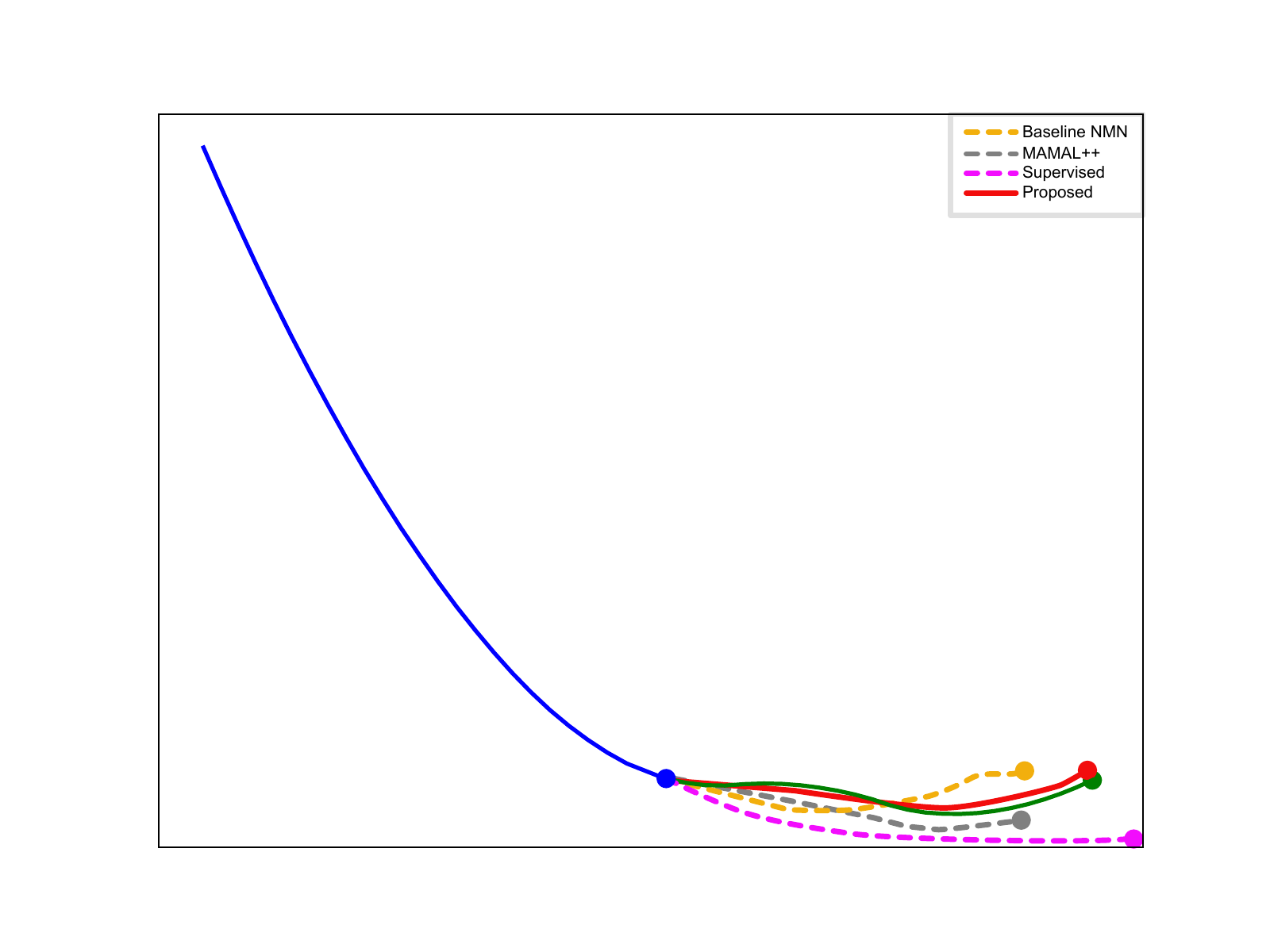}}
    \subfloat[][]{\includegraphics[width=.45\linewidth]{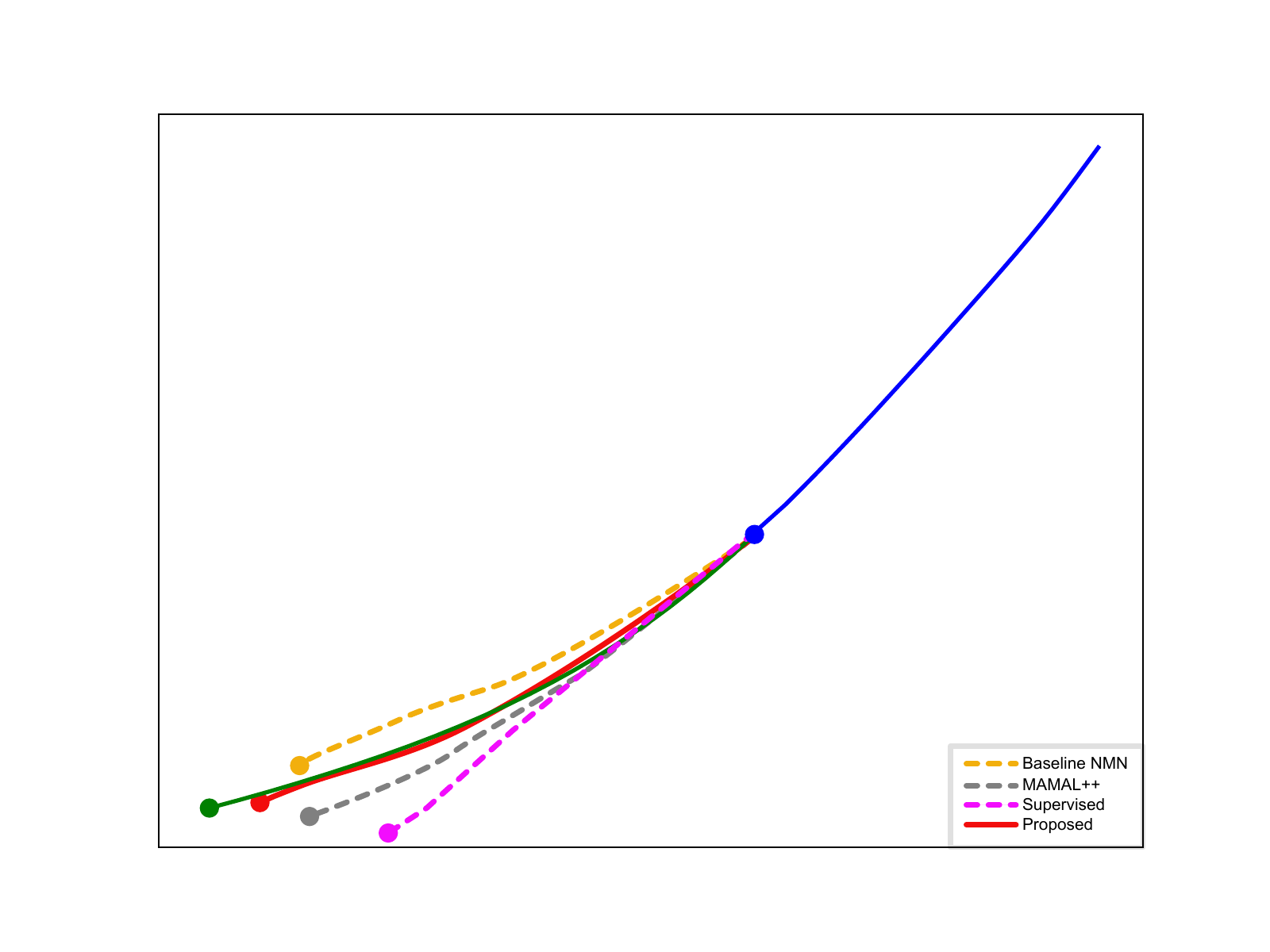}}\\
    \subfloat[][]{\includegraphics[width=.45\linewidth]{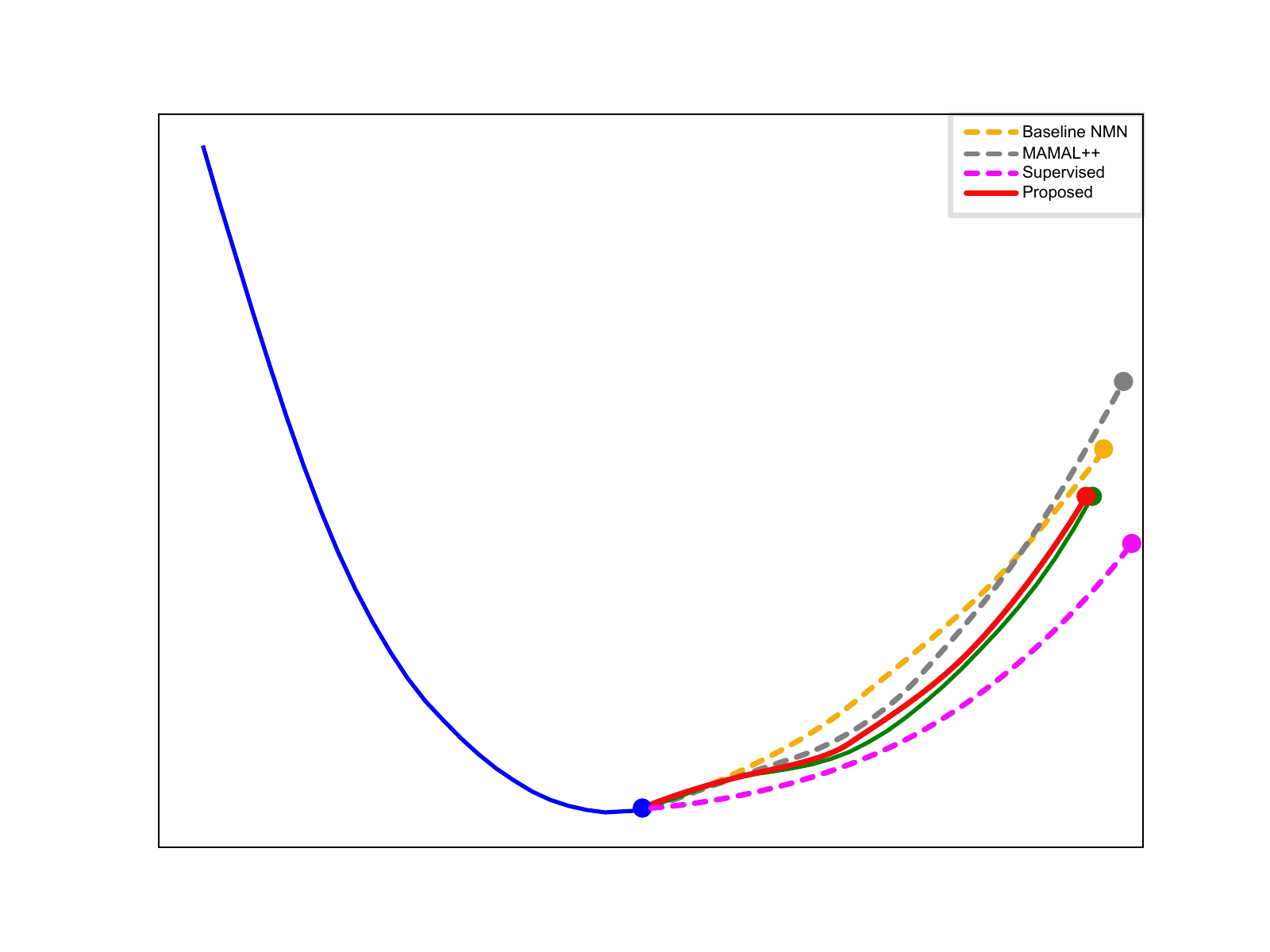}}
    \subfloat[][]{\includegraphics[width=.45\linewidth]{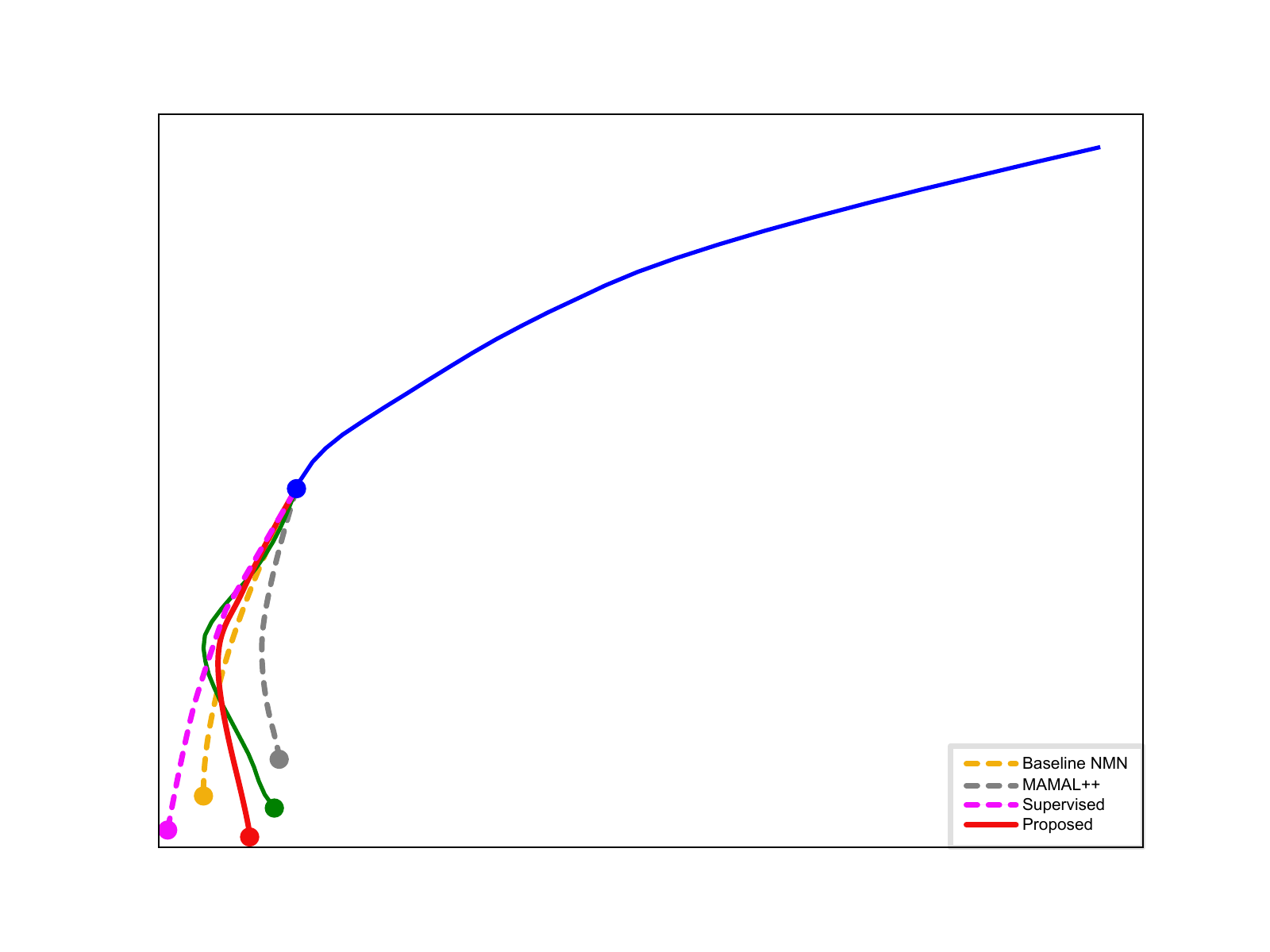}}
    \caption{Qualitative results for four randomly selected pedestrian trajectories. The observed part of the trajectory is shown in blue. The ground truth future trajectory is given in green. Predictions shown are as follows: the proposed method is in red; MAML++ \cite{antoniou2018train} is shown in gray; the supervised learning model of \cite{fernando2018gd} is in purple; and the baseline memory model is in orange.}
    \label{fig:track_qualitative}
\end{figure*}

Qualitative results for 4 randomly selected samples are shown in Fig. \ref{fig:track_qualitative}. Considering the challenging nature of the provided examples we conclude that the proposed method, with the aid of the external memory bank to store informative facts and an augmented memory access and consolidation scheme which assists knowledge retrieval, has more capacity to quickly adapt to the task at hand and better anticipate future behaviour. Furthermore, we observe that even with the fine-tuning process, the supervised learning \cite{fernando2018gd} baseline hasn't been able to grasp the irregular nature in the pedestrian trajectories compared to the cyclist trajectories (on which the models are trained on). The MAML++ \cite{antoniou2018train} and baseline NMN models have been able to generate more accurate predictions compared to the supervised learning baselines, however, both approaches are outperformed by the proposed model. 

When considering the quantitative and qualitative results it is clear that the proposed model has greater flexibility when adapting the learnt knowledge for the new task at hand. Furthermore, we observe that it continues to improve the predictions with additional training steps. However, with the baseline methods we observe that these methods fail to adapt to the new task, indicating that the back-propagation algorithm fails to update the controllers such that it learns the salient attributes from the limited available data. 

\section{Discussion}
In this section we provide evidence demonstrating the importance of the SG predictors in the memory controllers, and illustrate further functionality of SGs in-terms of knowledge sharing. 

\subsection{The Importance of SG Predictors in the Controllers}
\label{sec:importance_of_SGs}
We believe that the superior results of the proposed model is a result of the utilisation of SG predictors in the controllers. For example, in comparison to the baseline NMN approach we observe an increase of approximately 20\% in accuracy for the classification task, and a 35\% reduction in loss is observed for the regression task. The architectures are identical except for the SG predictors in the memory controller modules. 

As illustrated in Sec. \ref{sec:sgs}, the controller modules in the NMN are the primary governor for the information flow into and out of the memory. Hence, in a meta-learning situation the controller should posses the capacity to quickly adapt their functionality to the new task at hand. To illustrate this we conduct an additional ablation evaluation using the MNIST dataset \cite{lecun1998mnist} as follows: We use the same proposed and baseline NMN architectures used in Sec. \ref{sec:classification} (i.e for the Omniglot classification task) and consider a 10-way classification for the MNIST digits. However, for the first 500 iterations of the algorithm the models see only the first 5 classes of MNIST. At every 500th iteration we put in a batch of 10 images randomly picked from the next 5 classes. The test data is composed completely of unseen samples from this second set of 5 classes. 

\begin{figure*}[htbp]
    \centering
    \subfloat[][]{\includegraphics[width=.49\linewidth]{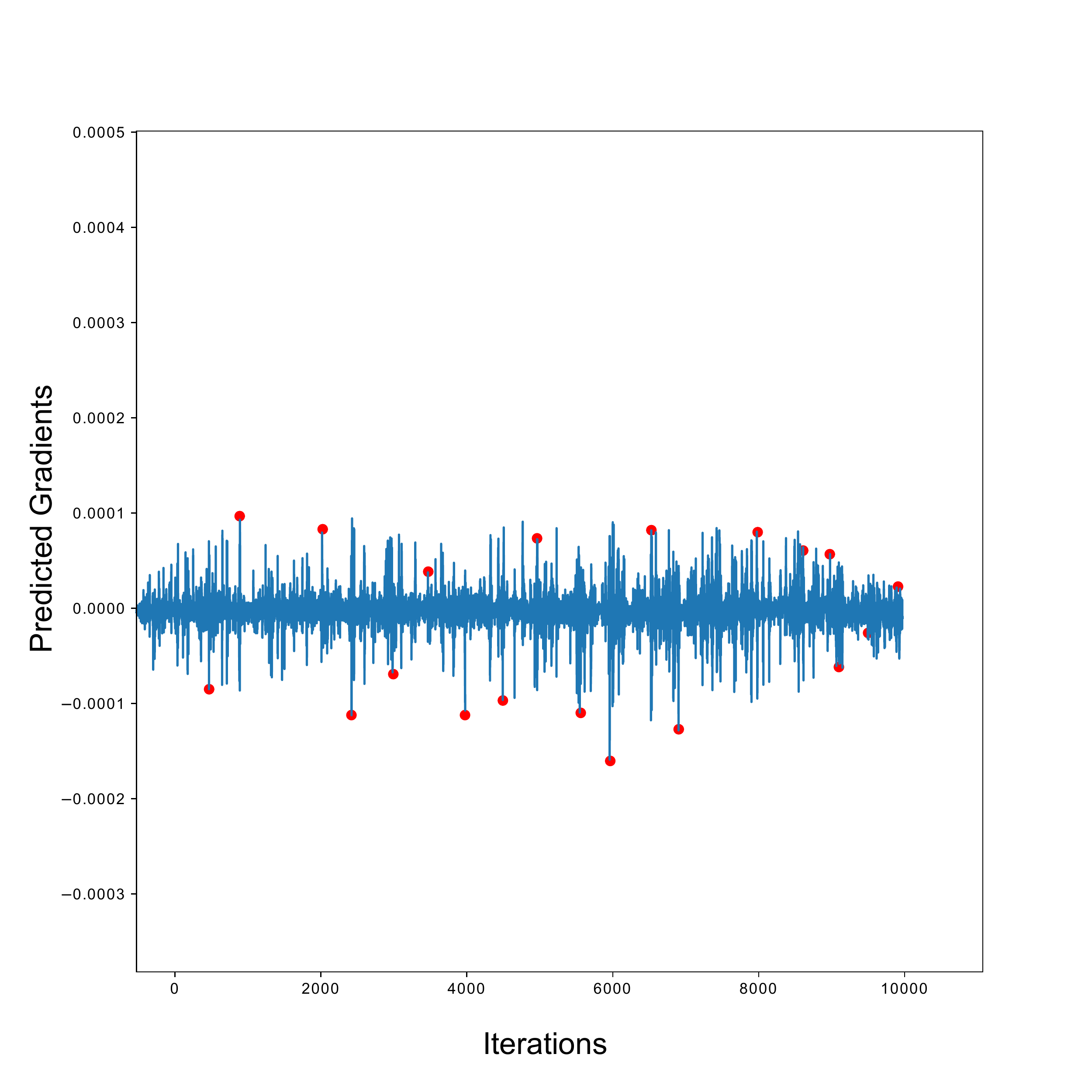}}
    \subfloat[][]{\includegraphics[width=.49\linewidth]{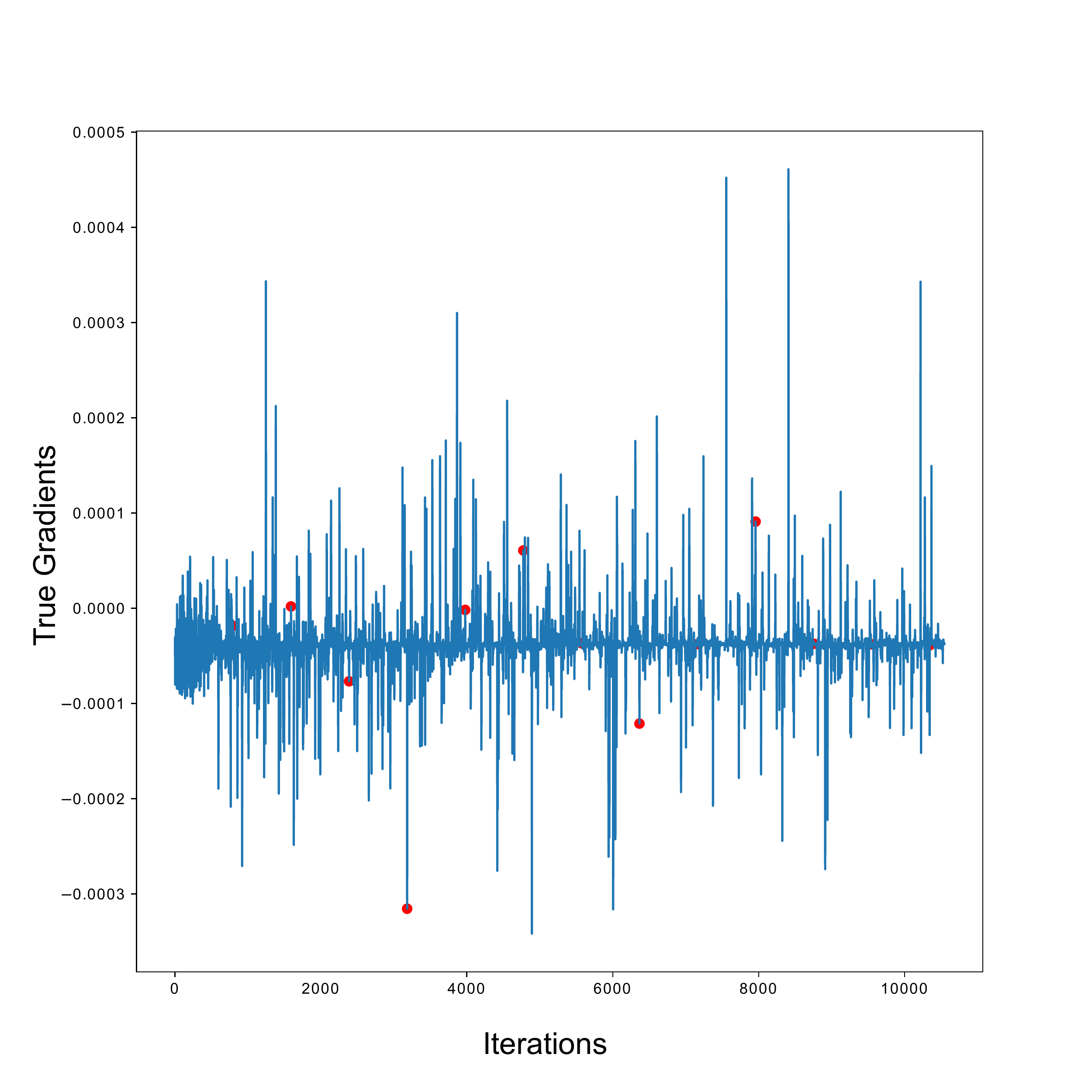}}
    \caption{Gradients generated by the SG predictor of the output controller (a) and through back-propagation (b) for the meta-testing evaluation using the MNIST dataset. The model performs 10-way classification, however, observes the second set of 5 classes of MNIST only once in every 500 iterations (indicated by the red dots).}
    \label{fig:mnist_grads}
\end{figure*}

\begin{figure*}[htbp]
    \centering
    \subfloat[][]{\includegraphics[width=.49\linewidth]{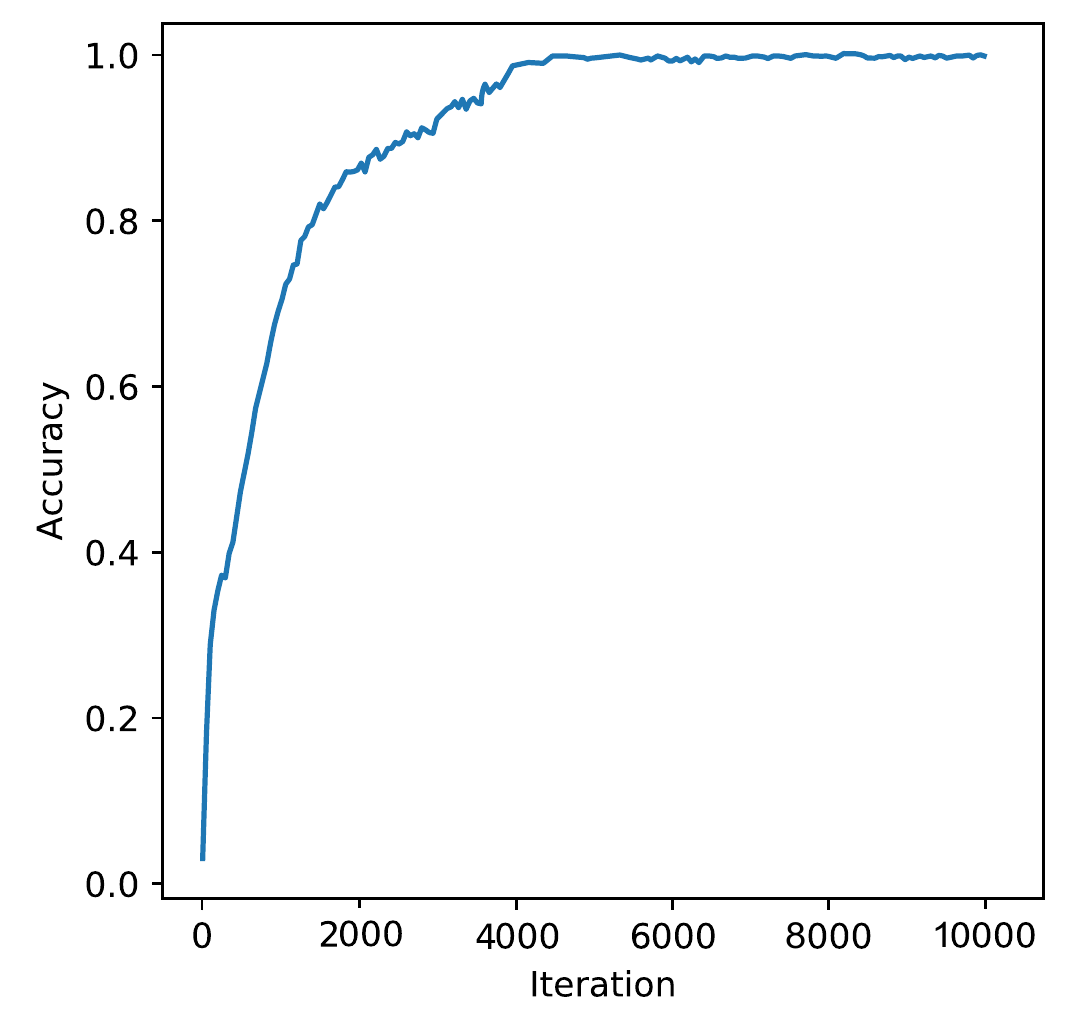}}
    \subfloat[][]{\includegraphics[width=.49\linewidth]{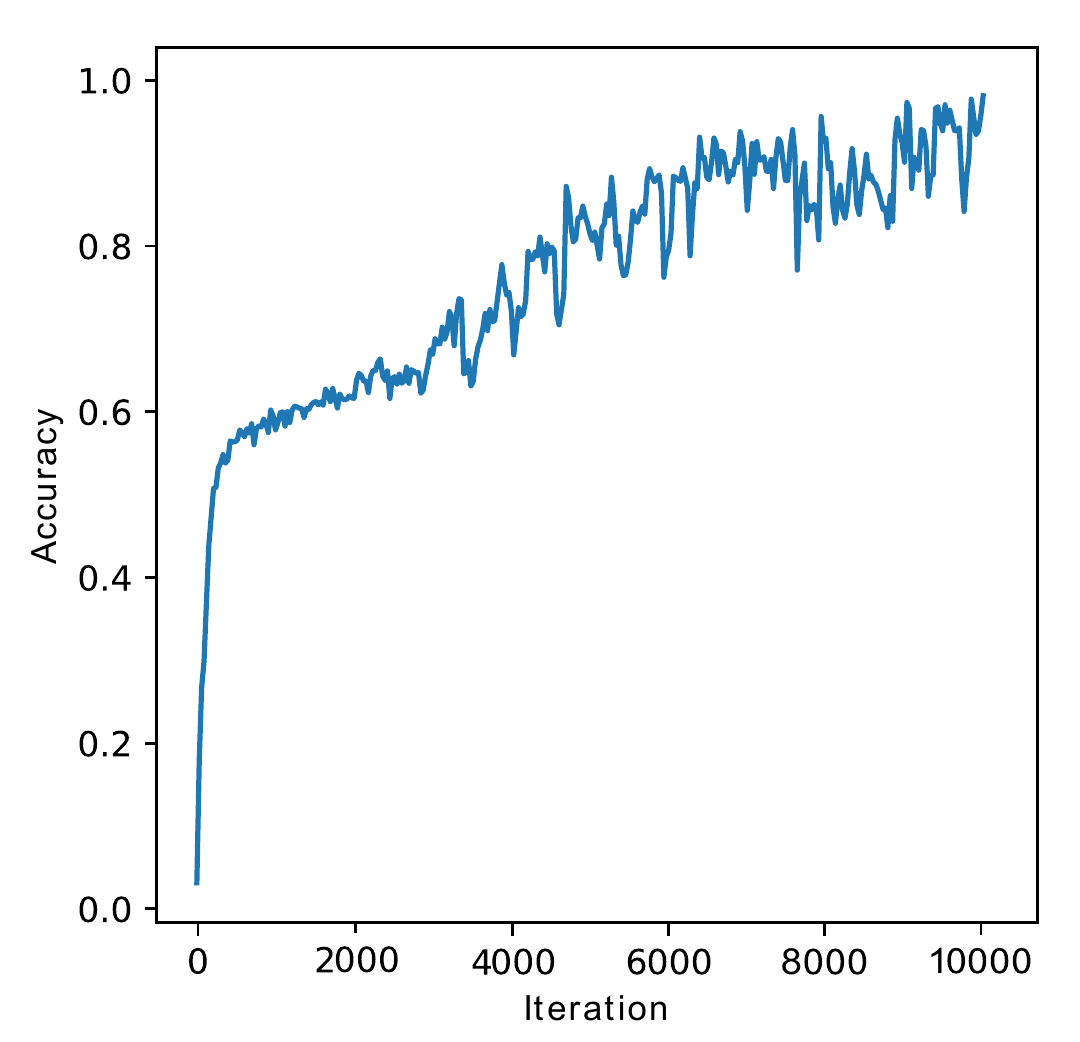}}
    \caption{Test accuracies for the proposed (a) and baseline (b) NMNs for the meta-testing evaluation using the MNIST dataset. The model performs 10-way classification, however, observes the second set of 5 classes of MNIST only once in every 500 iterations. Note that the test batch is composed of unseen examples from the second set of 5 classes.}
    \label{fig:mnist_learning_curves}
\end{figure*}

Fig. \ref{fig:mnist_grads} (a) and Fig. \ref{fig:mnist_grads} (b) illustrate how the error gradients from the output controller behave when training the proposed and baseline NMNs, respectively. Using red dots we have indicated the batches in which the images from the secondary classes were presented to the model. In Figs. \ref{fig:mnist_learning_curves} (a) and \ref{fig:mnist_learning_curves} (b) we provide the test accuracies for the respective models. 

When analysing the plots it is clear that gradients emulated by the SG predictors are more stable than the gradients generated through back-propagation. Most importantly, in Fig.\ref{fig:mnist_grads} (a) we observe that the peaks in the plot correspond to the meta-testing iterations, indicating that the controllers are quickly made to learn using the new information. In contrast, the large gradient updates for the baseline memory are for the primary training data, indicating that the information from the meta-learning task is discarded or over-written.  

\subsection{Knowledge Sharing with Controller SGs}

\begin{figure*}[htbp]
    \centering
    \subfloat[][Shared SGs]{\includegraphics[width=.95\linewidth]{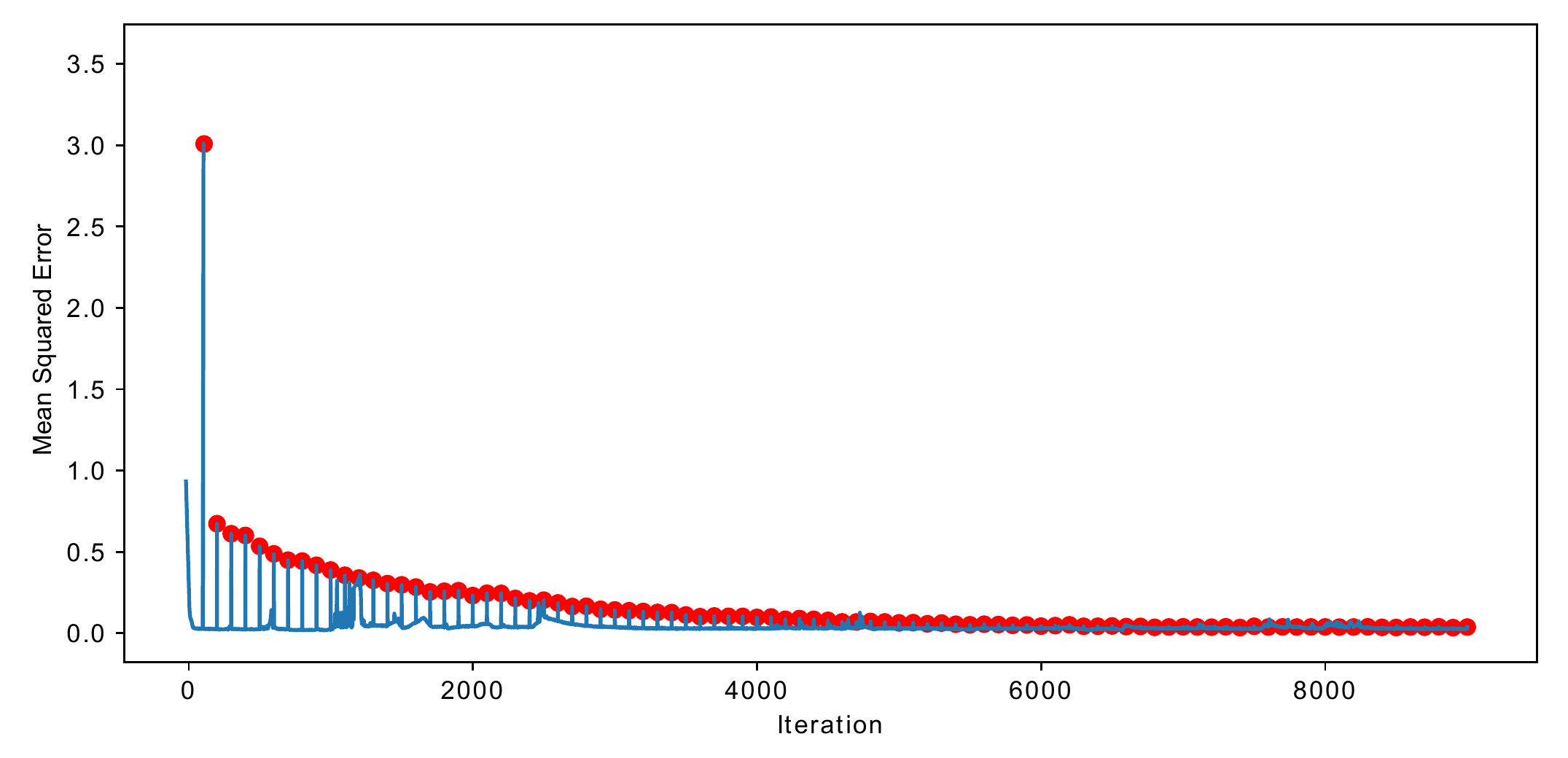}}\\
    \subfloat[][Separate SGs]{\includegraphics[width=.95\linewidth]{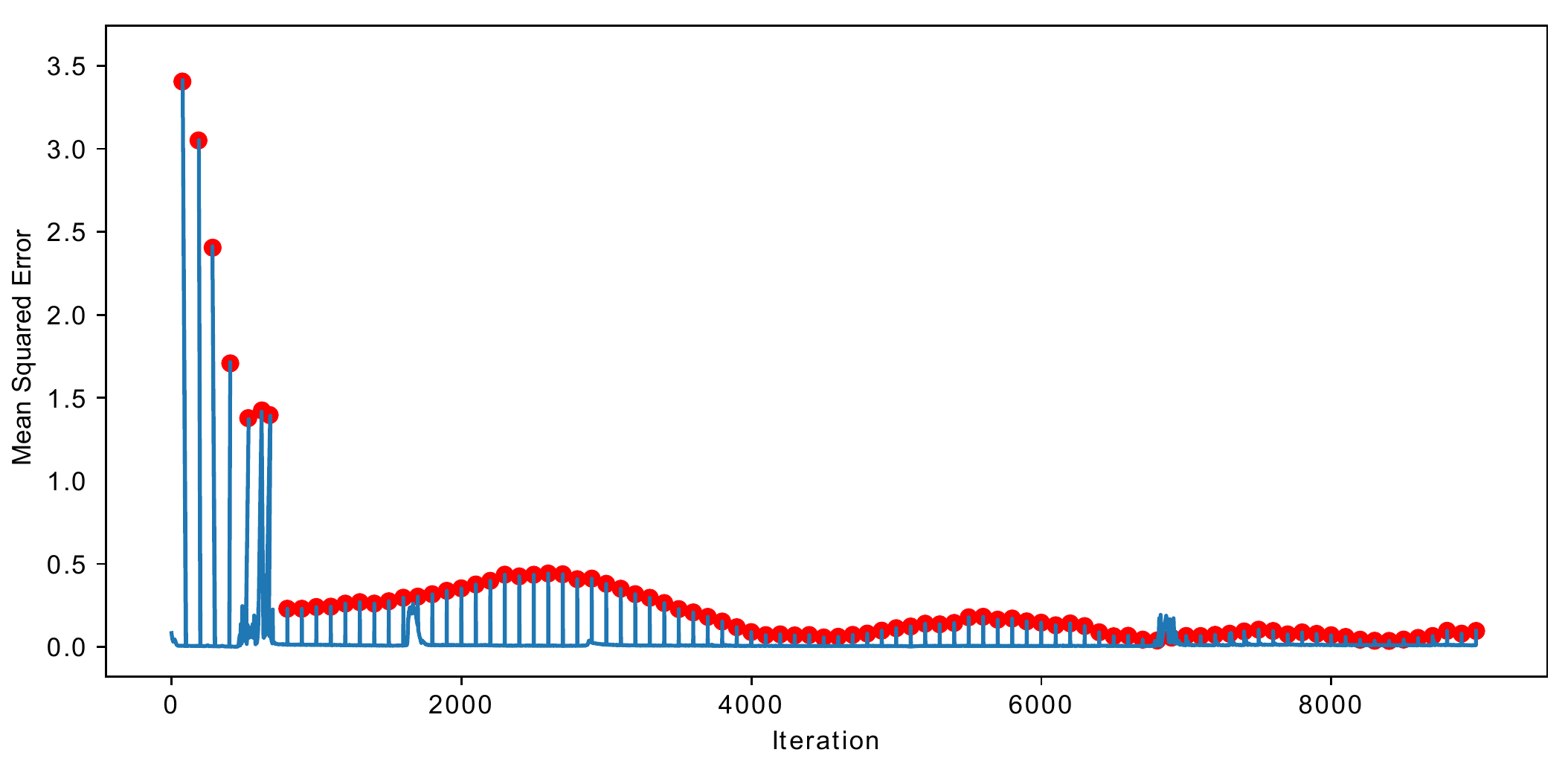}}
    \caption{Training curves for the knowledge sharing experiment with the inD trajectory prediction task. Two separate models using the proposed NMN architecture are used for the cyclists and pedestrians, respectively. Cyclist trajectories are fed to the cyclist stream as the primary task, however, the pedestrian data passes through the pedestrian model only once in every 100 iterations (denoted by a red circle).}
    \label{fig:knowledge_sharing}
\end{figure*}

As illustrated in Sec. \ref{sec:importance_of_SGs} the controllers play a vital role when extracting salient information from the historical data stored in memories, and the SG predictor modules can guide controllers to learn fast. To demonstrate the practical implications of these findings we conduct an additional experiment using the trajectory dataset used in Sec. \ref{sec:regression}. Similar to the previous evaluation we filtered out the trajectories from cyclists and pedestrians. In this experiment we use two instances of the proposed NMN architecture, with two $f^E$s, two memory stacks, two sets of controllers and two $f^D$s, for the two trajectory types. However, the SG predictors are shared among the controller modules. As the baseline we used the same architecture where the SG predictors are also separate. For the first 100 iterations we feed the data to the cyclist stream and subsequently at every 100th iteration 10 examples from the pedestrian data are fed to the pedestrian branch. It should be noted that except in these iterations, no updates were made to the pedestrian model. Figs. \ref{fig:knowledge_sharing} (a) and \ref{fig:knowledge_sharing} (b) illustrate the learning curves for the shared and separate SG models. The red dots indicate the iterations where pedestrian trajectories were fed to the pedestrian model. The results indicate that we can quickly train a secondary memory model from scratch to perform a novel task by sharing the gradient predictors. As the SG predictors are aware of the salient components in the data that they should consider when performing the trajectory prediction, a complete secondary model (i.e a $f^E$, memory stacks, controllers, and a $f^D$) can be quickly learned, even from scratch, with a relatively small amount of data and few iterations. This is a valuable trait for instances such as deploying a framework in a new environment where hand annotated data is scarce.  

\begin{figure*}[htbp]
    \centering
    \subfloat[][]{\includegraphics[width=.30\linewidth]{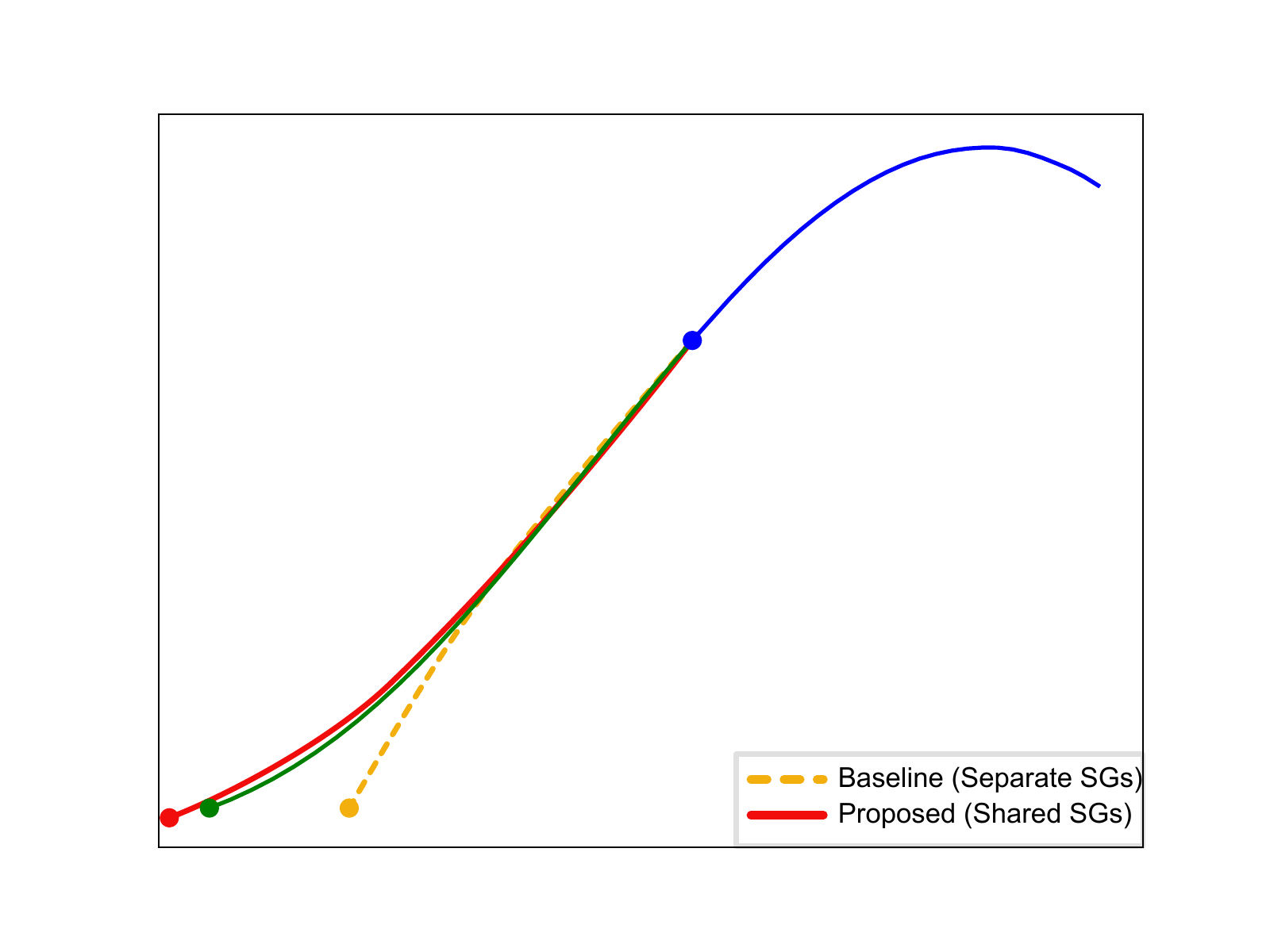}}
    \subfloat[][]{\includegraphics[width=.30\linewidth]{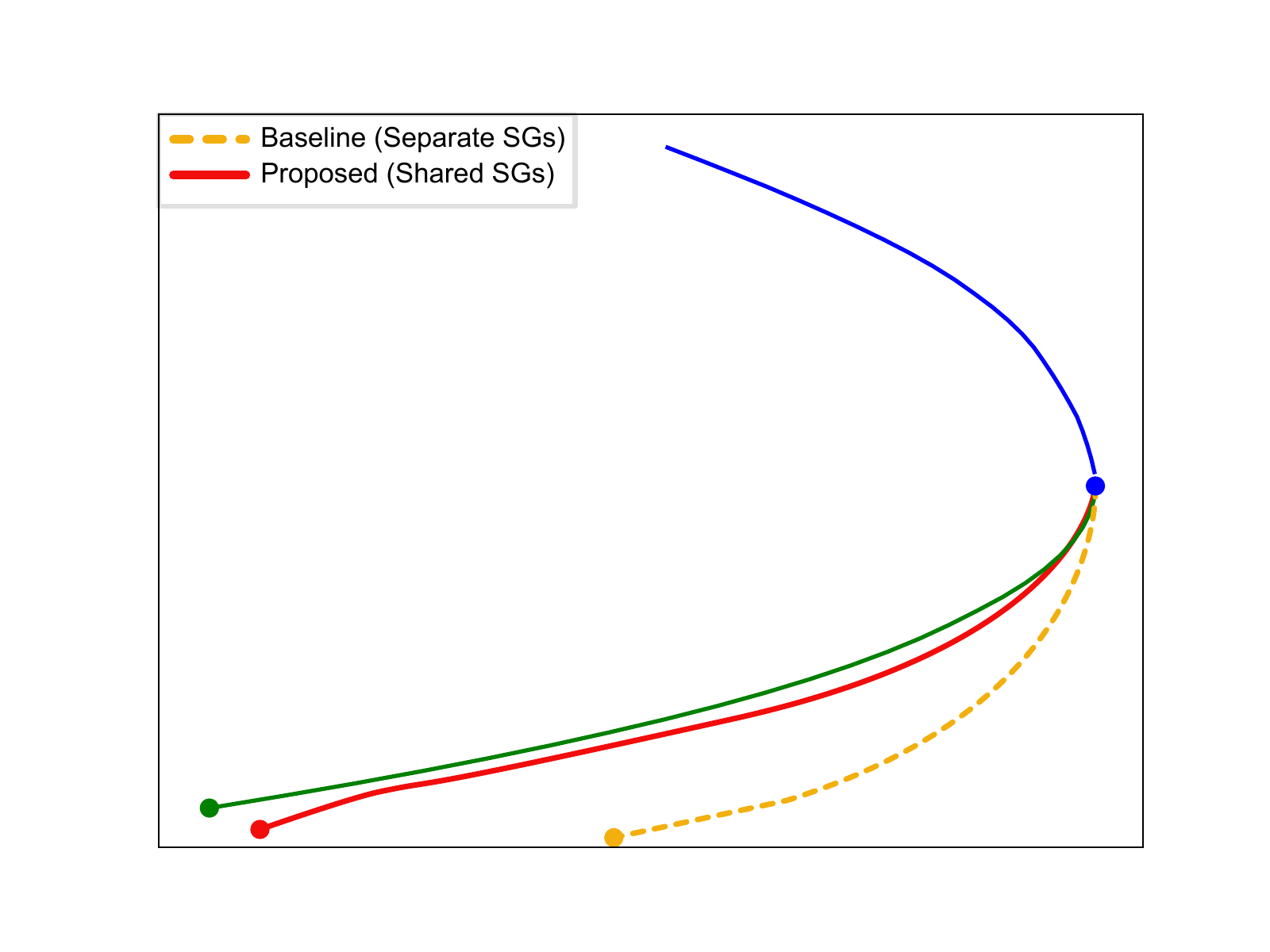}}
    \subfloat[][]{\includegraphics[width=.30\linewidth]{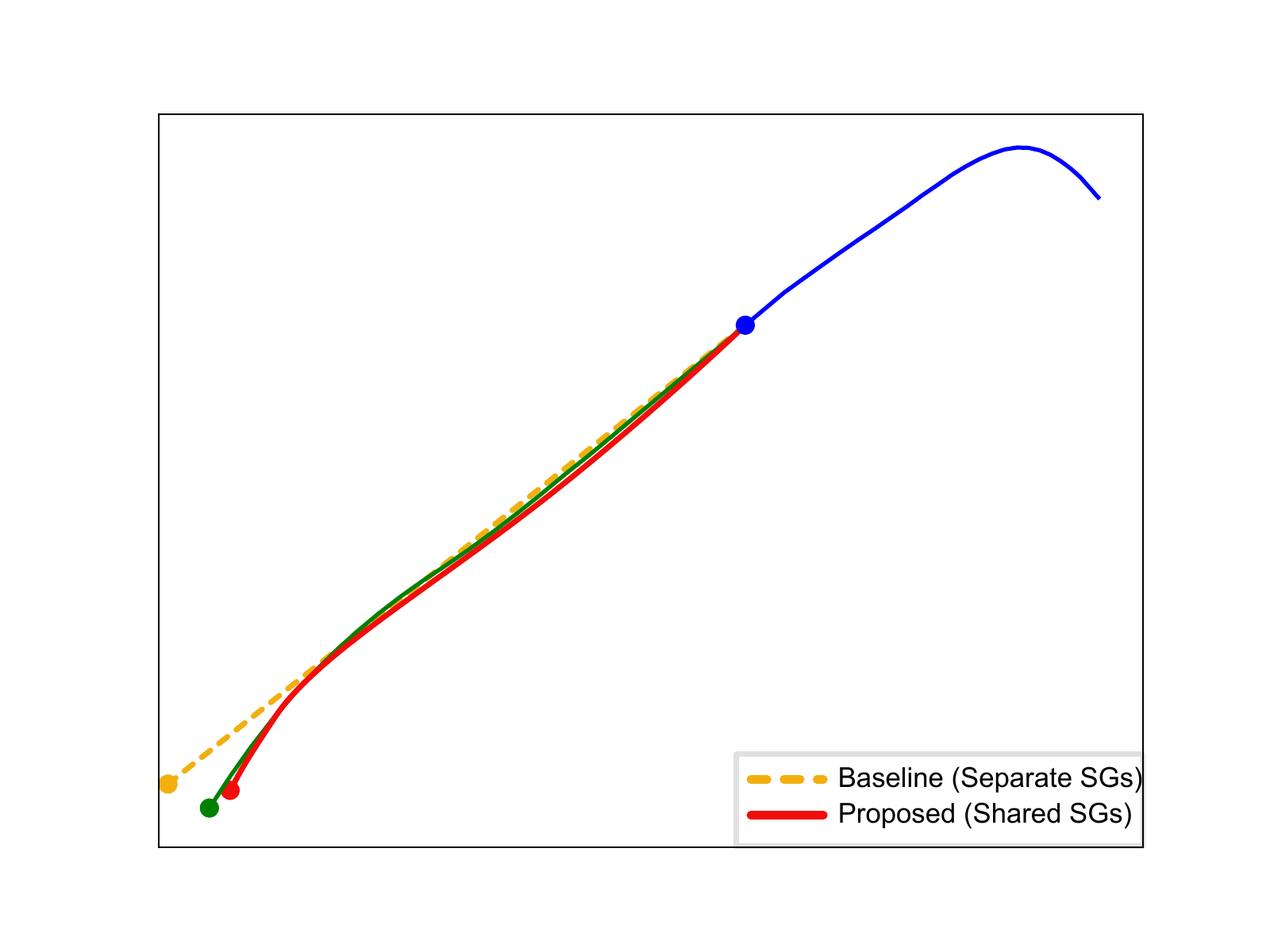}}\\
    \subfloat[][]{\includegraphics[width=.30\linewidth]{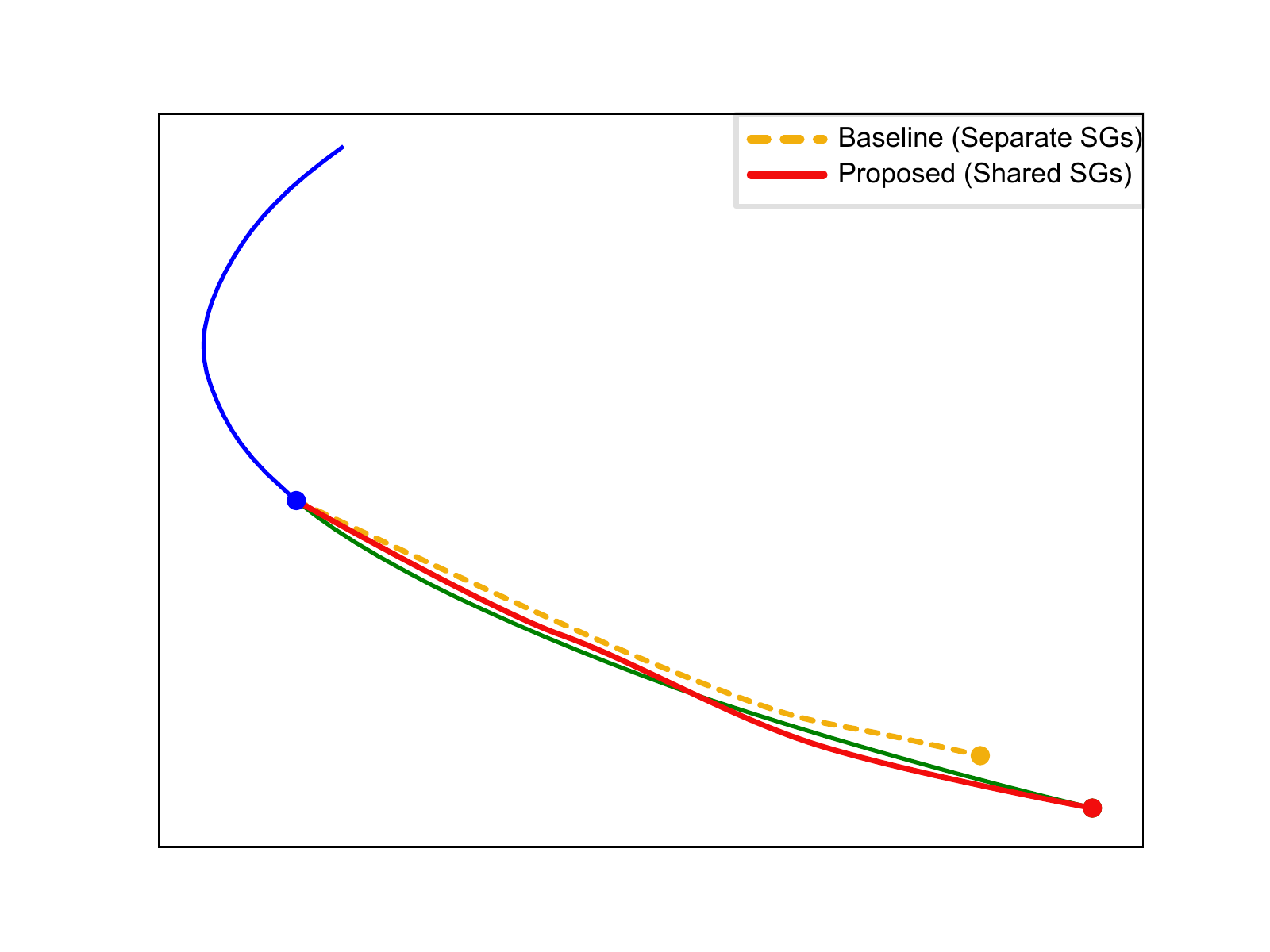}}
    \subfloat[][]{\includegraphics[width=.30\linewidth]{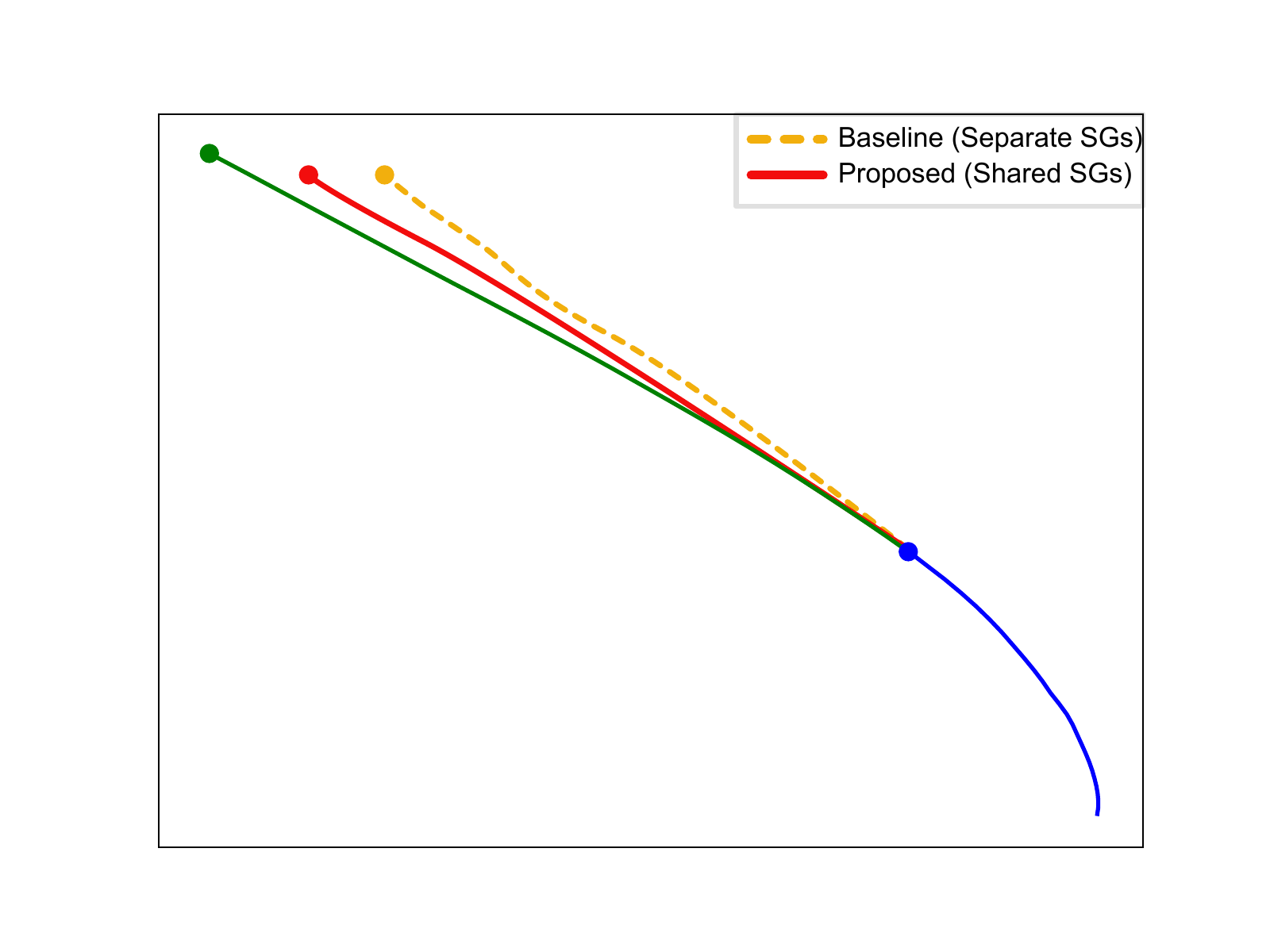}}
    \subfloat[][]{\includegraphics[width=.30\linewidth]{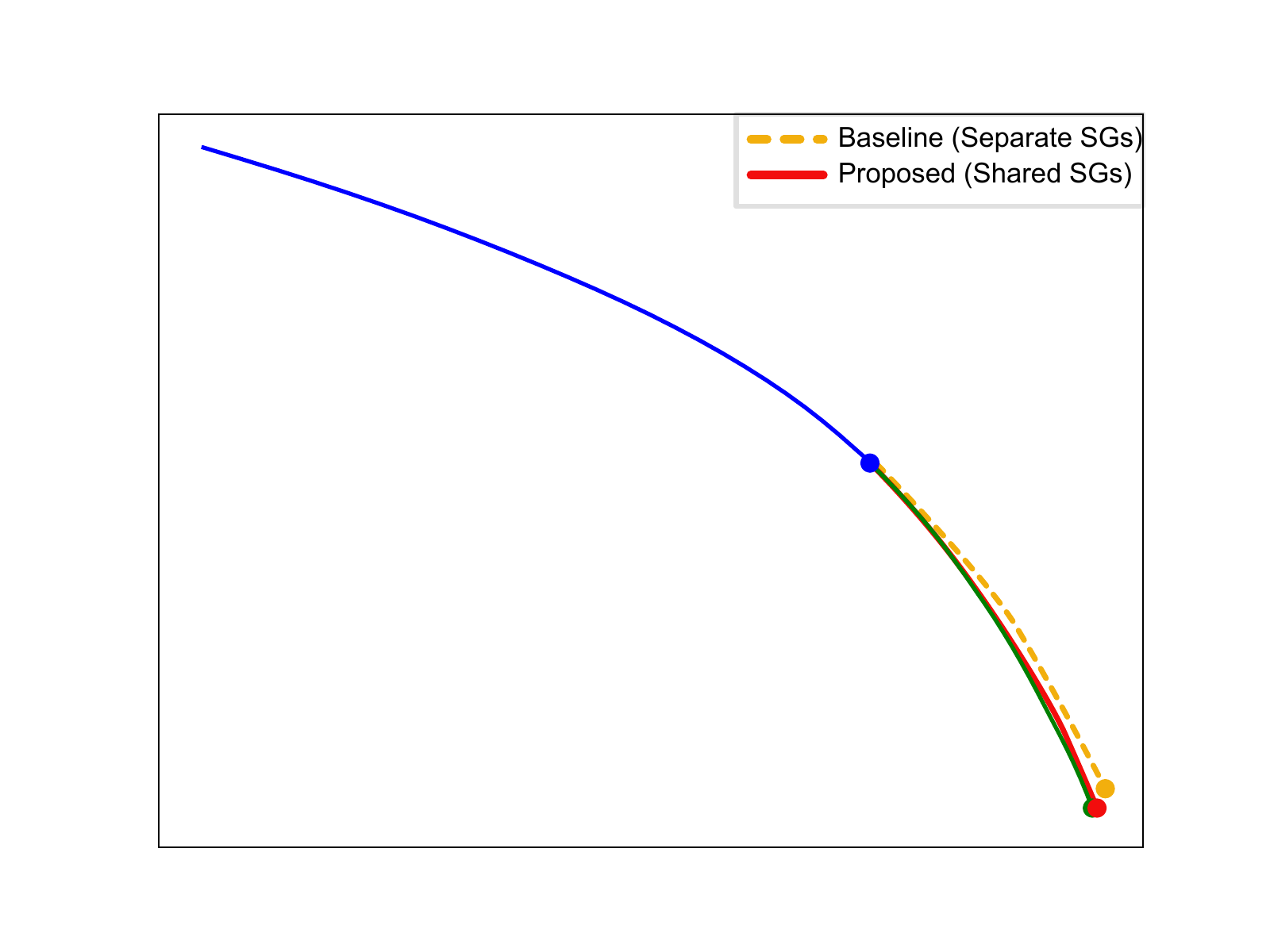}}
    \caption{Qualitative results for the knowledge sharing experiment. The observed part of the trajectory is shown in blue. The ground truth future trajectory is given in green. Predictions from proposed shared SG controller module are shown in red; and predictions from the separate SG controller module are shown in orange. (a)-(c) show predictions for cyclist trajectories, and (d)-(f) show predictions for pedestrian trajectories.}
    \label{fig:knowledge_sharing_qualitative}
\end{figure*}

Qualitative results of this experiment are given in Fig. \ref{fig:knowledge_sharing_qualitative}. In the first row we provide predictions from the cyclist stream where the input is a bicycle trajectory, and the second row shows the predictions for the pedestrian trajectories (i.e input is a pedestrian trajectory). We also provide predictions for when these individual streams have separate SG modules. An interesting observation from the presented qualitative results is that the proposed shared SG model is able to transfer the knowledge acquired when modelling pedestrian trajectories to the cyclist trajectory modelling task and model irregular cyclist trajectories. For example, in Fig. \ref{fig:knowledge_sharing_qualitative} (c) we observe an irregular turn towards the end of the future trajectory segment and the proposed shared SG method has been able to successfully anticipate this behaviour. Moreover, we observe that this knowledge sharing does not cause adverse effects and diminish the ability to capture characteristics which are specific to the individual streams.  

\section{Conclusion}
In this paper, we propose a novel controller learning mechanism for Neural Memory Networks (NMNs) which exploits advances in NMNs and synthetic gradient based model de-coupling. The introduced de-coupling process allows the NMN controllers to formulate arbitrary feedback paths and demonstrate rapid adaptation of the stored knowledge in the presence of new information. Through the evaluation conducted on challenging meta-learning tasks, which includes both classification and regression scenarios, we demonstrate that our proposed memory architecture is able to outperform all considered baselines. By carefully analysing the temporal evolution of the error gradients of the baseline and proposed methods, we illustrate the power of our controller learning strategy to capture salient information cues that are needed for meta-learning tasks. In addition, we demonstrate how these synthetic feedback paths can be utilised as a knowledge sharing mechanism with NMNs. In future work, we will explore applications of the proposed innovations for aligning unsynchronised multi-modal input streams via the proposed knowledge sharing mechanism, and summarise the relationships among the learned abstract representations. 

\section*{References}

\bibliography{egdb.bib}

\end{document}